\documentclass[11pt]{article}

\usepackage[final]{acl}

\usepackage{times}
\usepackage{latexsym}

\usepackage[T1]{fontenc}
\usepackage[utf8]{inputenc}

\usepackage{microtype}

\usepackage{inconsolata}

\usepackage{graphicx}

\usepackage{enumitem}

\usepackage{booktabs}  % 예쁜 테이블 선 (\toprule, \midrule 등)
\usepackage{amssymb}   % 체크마크 (\checkmark) 사용
\usepackage{multirow}  % 셀 병합
\usepackage{graphicx}  % 표 크기 조절 (\resizebox)
\usepackage{xcolor}
\usepackage{array}
\usepackage{tcolorbox} % prompt foramt용
\usepackage[T1]{fontenc}
\usepackage{listings}
\usepackage{amsmath}
\usepackage{amssymb}
\usepackage{mathtools}
\usepackage{amsthm}
\tcbuselibrary{listings,breakable}
\lstdefinestyle{promptstyle}{
  basicstyle=\small\ttfamily,  % \ttfamily\small
  breaklines=true,
  breakatwhitespace=true,
  columns=fullflexible,
  keepspaces=true
}

\definecolor{errorred}{RGB}{200, 0, 0}
\definecolor{correctblue}{RGB}{0, 0, 150}
\definecolor{hintgreen}{RGB}{0, 100, 0}
\definecolor{NavyBlue}{RGB}{8,111,189}

\title{DCGC: Draft-Conditioned Global Correction for\\ Complex Reasoning with Masked Diffusion Models}

\author{
 \textbf{Minhae Oh\textsuperscript{1,*}},
 \textbf{Nakyung Lee\textsuperscript{1,*}},
 \textbf{Jungwoo Lee\textsuperscript{1,\textdagger}}
 \\
 \textsuperscript{1}Department of Electrical and Computer Engineering, Seoul National University 
 }

\begin{document}

\maketitle

\begin{abstract}
Correcting flawed reasoning traces remains a significant challenge for Large Language Models (LLMs), whose autoregressive generation can propagate early mistakes into subsequent reasoning. 
We introduce \textbf{DCGC}, a Masked Diffusion Model (MDM) framework for global correction that uses an imperfect solution draft from an upstream solver as auxiliary context. 
DCGC combines task-specific Supervised Fine-Tuning (SFT) with a novel inference-time mechanism called \textbf{Dynamic Dual-CFG}. This mechanism separates problem-only and joint problem-draft branches and scales the draft-conditioned residual using a relative confidence gap.
Across math, code, and knowledge reasoning benchmarks, DCGC outperforms standard sampling and simpler CFG variants, with additional results suggesting transfer to different diffusion backbones. In test-time setting where ground-truth failure labels are unavailable, DCGC improves full test set accuracy by correcting low-consensus upstream outputs, highlighting its utility as a verifier-free global correction module for difficult reasoning instances.
\end{abstract}
% We introduce \textbf{DCGC}, a Masked Diffusion Model (MDM) framework that uses an imperfect solution draft generated by an upstream solver as auxiliary.

\section{Introduction}

While LLMs have recently demonstrated remarkable performance on complex reasoning tasks, they remain prone to logical fallacies and hallucinations, particularly in multi-step reasoning settings~\cite{cot,hallucination,logicalfallacies}. As a result, numerous studies have investigated self-refinement, where models iteratively critique and revise their own generations to improve solution quality~\cite{improving}. Prominent approaches such as Self-Refine~\cite{selfrefine}, Reflexion~\cite{reflexion}, and LATS~\cite{lats} have been proposed to mitigate these issues by incorporating feedback loops or search mechanisms. However, since the predominant architecture of contemporary LLMs is autoregressive, most existing self-refinement approaches are built upon this strictly sequential, left-to-right paradigm. This makes correction difficult, as an early mistaken step in the revised trajectory becomes part of the prefix that conditions subsequent tokens, which may then continue along the same erroneous path~\cite{snowball,limitation}.
Moreover, recent studies have shown that Autoregressive Models (ARMs) are often poorly calibrated and exhibit overconfidence in their own predictions~\cite{taming,lmknow,calibration}. Such overconfidence can make tool-free self-correction brittle, since the model may over-trust its own reasoning trajectory even when it contains errors~\cite{refinebench,dontknow}. 

This motivates considering Diffusion Language Models (DLMs), particularly MDMs, as a structurally distinct substrate for post-hoc reasoning correction~\cite{diffuseq,diffimprove}. Unlike ARMs that commit to a single left-to-right prefix, MDMs generate text through iterative denoising over masked positions, allowing the output sequence to be revisited during inference. This property makes MDMs well suited for \emph{draft conditioned global correction}, where the model uses an imperfect draft as auxiliary context during global denoising while remaining strongly conditioned on the original problem. Despite this potential, prior work has primarily utilized MDMs for generating solutions from scratch, leaving their use for global correction of long reasoning traces underexplored.

In this work, we introduce \textbf{DCGC} (\textbf{D}raft-\textbf{C}onditioned \textbf{G}lobal \textbf{C}orrection with Masked Diffusion Models), a framework that uses MDMs as draft-aware global correction modules for complex reasoning. Given a problem and an imperfect draft generated by an upstream solver, DCGC performs iterative denoising under both problem-only and joint problem-draft contexts. The draft influences generation through the joint context, while its additional residual contribution is controlled relative to the problem-only branch. Unlike methods that rely on external tools, memory buffers, or computationally expensive tree searches~\cite{tot}, DCGC operates in a tool-free setting using only internal model signals.

To realize this, DCGC combines dual-capability SFT with Dynamic Dual-CFG. The SFT stage trains the MDM on both problem-only solving and draft-conditioned correction, equipping the model with both conditioning modes used during inference. Dynamic Dual-CFG then separates problem-only and joint problem-draft branches, using their relative confidence gap to modulate draft influence during denoising. This design encourages selective draft reuse without treating internal confidence as a verifier of correctness. 

Our contributions are summarized as follows:
\begin{itemize}[itemsep=0pt, topsep=0pt, parsep=0pt, partopsep=0pt, leftmargin=*]
    \item We propose \textbf{DCGC}, a draft-conditioned global correction framework that repurposes MDMs as tool-free correction modules for complex reasoning. DCGC treats an imperfect solution as auxiliary context for global denoising and correction.

    \item We introduce \textbf{Dynamic Dual-CFG}, an inference-time guidance mechanism that separates problem-only and draft-conditioned branches. By scaling draft guidance with a relative confidence gap, the method controls how strongly the joint context contributes beyond the problem-only branch.
    
    \item We provide controlled empirical evidence across math, code, and knowledge reasoning benchmarks. 
    % DCGC achieves the highest average accuracy of 24.8\% on solver-failure hard sets, outperforming the strong baselines.
    DCGC improves initially failed solutions on solver-failure hard sets, achieving 24.8\% average accuracy over strong autoregressive and diffusion-based baselines.
\end{itemize}

\section{Related Works}
\paragraph{Non-Autoregressive Generation and Editing.} While standard ARMs attempt iterative reasoning refinement via prompting or tool use~\cite{selfrefine, reflexion, lats, improving}, they inherently suffer from sequential error propagation and lack global editing flexibility~\cite{refinebench, selfincorrect, cannotself}. Recently, DLMs have emerged as parallel, non-autoregressive alternatives~\cite{d3pm, sedd, mdlm, llada, mercury}, yet they are predominantly utilized for \textit{de novo} generation rather than refinement~\cite{surveydlm, beyond}. Bridging this gap, our framework repurposes DLMs as specialized reasoning refiners. Inspired by visual editing techniques~\cite{instructpix2pix}, we leverage CFG to steer the global denoising process, enabling non-sequential, fine-grained correction of flawed initial solutions while preserving valid logic.
% Numerous approaches have been developed to enhance the reasoning capabilities of LLMs through iterative refinement, employing strategies that range from inference-time prompting to tool-using~\cite{selfrefine, reflexion, lats, improving}. However, these methods are predominantly built upon ARMs, which inherently suffer from error propagation and lack the flexibility to perform non-sequential edits efficiently~\cite{refinebench, selfincorrect, cannotself}.

% Recently, DLMs have emerged as a compelling non-autoregressive alternative to traditional Transformer-based architectures~\cite{d3pm, sedd, mdlm, llada, mercury}. By modeling generation as a parallel denoising process, DLMs facilitate global exploration and diverse output generation, offering significant potential for complex reasoning tasks~\cite{surveydlm, beyond}. However, these approaches primarily utilize diffusion for \textit{de novo} generation. There has been no significant attempt to exploit the intrinsic editability of DLMs as a specialized \textit{refiner} tool, specifically designed to intervene and correct intermediate errors within a long reasoning chain. Our framework bridges this gap by repurposing the diffusion mechanism for refinement. Drawing inspiration from visual editing methods such as InstructPix2Pix~\cite{instructpix2pix}, we employ CFG to steer the denoising process, enabling fine-grained correction of LLM outputs while preserving valid logic. 

\paragraph{Adaptive Scaling in CFG.} To overcome the limitations of static CFG, recent methods dynamically modulate guidance scales across spatial regions~\cite{scfg}, temporal denoising steps~\cite{betacfg}, or internal model confidence~\cite{insitu}. However, these approaches are designed for single-condition trajectories and lack the mechanisms to filter misleading auxiliary signals in multi-context scenarios. In contrast, we propose a differential confidence strategy that modulates guidance based on the relative certainty gain from the secondary context, allowing the refiner to selectively use auxiliary information only when it improves confidence.
% Recent studies have proposed dynamic scaling strategies to overcome the limitations of static CFG. S-CFG \cite{scfg} adaptively modulates guidance scales across different semantic regions to enhance local fidelity and prevent over-exposure. To optimize temporal dynamics, $\beta$-CFG \cite{betacfg} introduces a time-dependent scheduler that adjusts guidance intensity throughout the denoising process. Furthermore, In-situ Autoguidance \cite{insitu} utilizes the model’s internal confidence to elicit self-correction adaptively during inference. However, these approaches primarily focus on single-condition trajectories and lack mechanisms to filter harmful auxiliary signals in multi-context scenarios. Instead, we propose a differential confidence strategy that modulates guidance based on the relative certainty gain provided by the secondary context.

\begin{figure*}[t!]
    \centering
    \includegraphics[width=\textwidth]{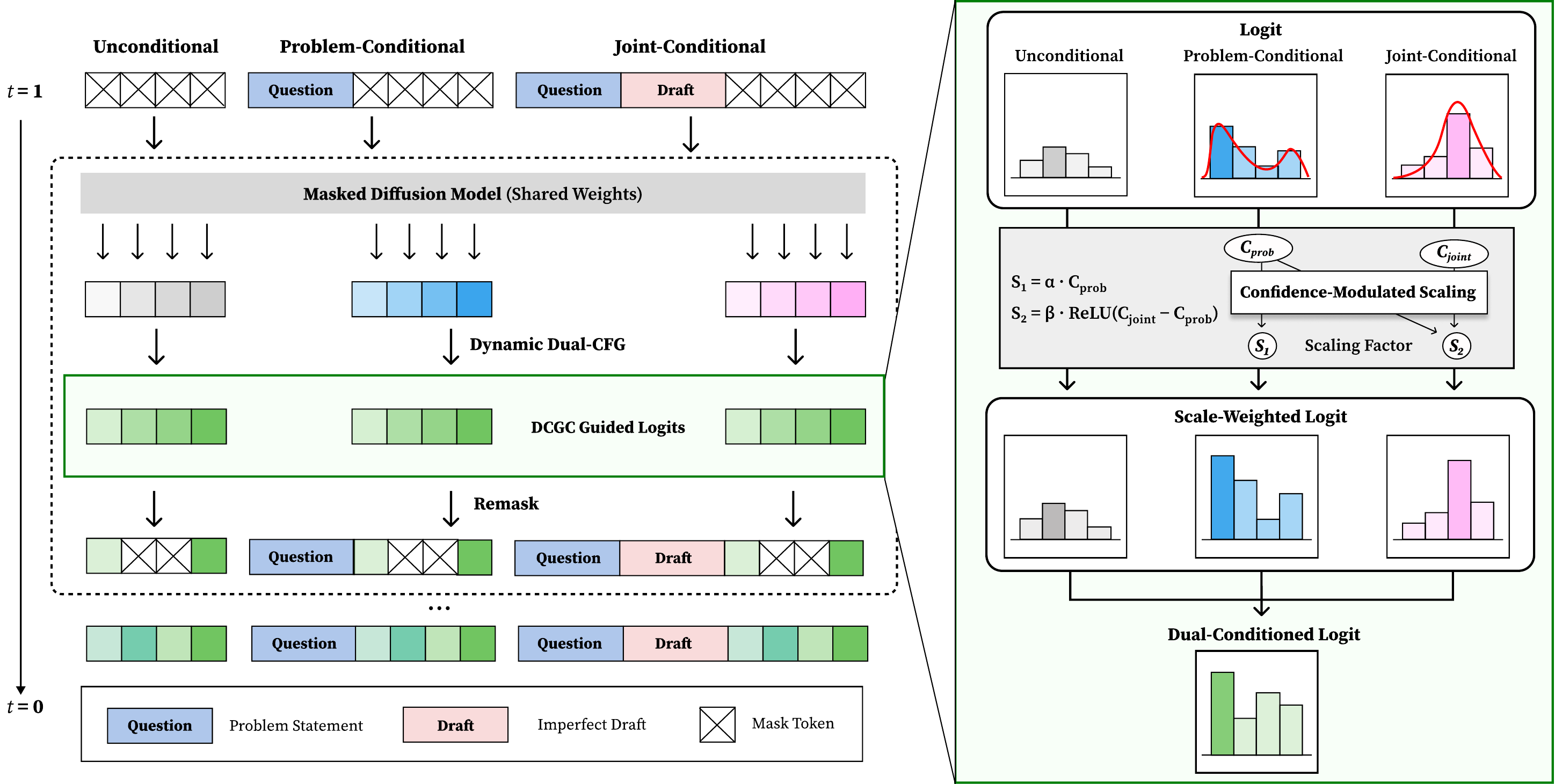}
    \caption{\textbf{Overview of DCGC.}  DCGC operates as an iterative denoising process (left) using three parallel conditioning streams at each step: unconditional, problem-conditioned ($Q$), and joint-conditioned ($Q, W$). The core Dynamic Dual-CFG mechanism (right) dynamically modulates guidance based on token-level confidence. By applying ReLU-based scaling ($S_2$), DCGC modulates the influence, activating the residual amplification only when the joint context provides a confidence gain($C_{joint} > C_{prob}$), ensuring robust selective reuse.}
    \label{fig:DCGC}
\end{figure*}

\section{Preliminaries}

\paragraph{Masked Diffusion Model.}
MDMs generate text via an iterative denoising process. Let $x_0=(x_0^1,\dots,x_0^L)$ be a sequence of clean tokens, and $x_t$ denote the corrupted sequence at timestep $t$.
The forward process $q(x_t \mid x_0)$ independently corrupts each token with a mask token $\text{[M]}$ according to a monotonically decreasing noise schedule $\alpha_t \in [0, 1]$:
\begin{equation}
\begin{aligned}
q(x_t \mid x_0)
&= \prod_{i=1}^L \Big(
    \alpha_t \, \delta(x_t^i, x_0^i) \\
&\qquad\quad
    + (1-\alpha_t) \, \delta(x_t^i, \text{[M]})
\Big)
\end{aligned}
\label{eq:forward_process}
\end{equation}
where $\delta(\cdot, \cdot)$ is the Kronecker delta function.
% The forward process $q(x_t \mid x_0)$ gradually corrupts $x_0$ by replacing tokens with a mask token $M$ according to a noise schedule $\alpha_t$ that decreases from 1 to 0.
% Specifically, each token is masked independently with probability $1 - \alpha_t$ at time $t \in [0, 1]$ as:
% \begin{equation} 
% q(x_t \mid x_0)= \prod_{i=1}^L \mathrm{Cat}\!\left(x_t^i;\ \alpha_t\,\delta_{x_0^i} + (1-\alpha_t)\,\delta_M
% \right).
% \label{eq:forward_process}
% \end{equation}
% Here, $\mathrm{Cat}(x;\pi)$ denotes a categorical distribution over the vocabulary (including $M$) with probability vector $\pi$, and $\delta_{x_0^i}$ and $\delta_M$ denote point-mass distributions at token $x_0^i$ and the mask token $M$, respectively.

The reverse process reconstructs $x_0$ from the corrupted state $x_t$ using a neural denoiser $\hat{x}_\theta(x_t, t)$~\cite{mdlm}. For any $0 \le s < t \le 1$, we define the reverse transition by posterior substitution as
\begin{equation}
p_\theta(x_s \mid x_t) ~\triangleq~ q\!\big(x_s \mid x_t,\ x_0=\hat{x}_\theta(x_t,t)\big).
\label{eq:reverse_subs}
\end{equation}

\paragraph{Training objective.}
Following LLaDA~\cite{llada}, we train the model to minimize the variational lower bound, which simplifies to the cross-entropy loss on masked positions:
\begin{equation}
\mathcal{L}(\theta)
~\triangleq~
-\mathbb{E}_{t,\,x_0,\,x_t}\!\left[
\frac{1}{|\mathcal{M}_t|}
\sum_{i \in \mathcal{M}_t}
\log p_\theta(x_0^i \mid x_t)
\right],
\label{eq:training_loss}
\end{equation}
where $\mathcal{M}_t = \{i \mid x_t^i = M\}$ denotes the set of indices masked at timestep $t$.

\paragraph{Classifier-Free Guidance (CFG).}
% Let $c$ denote the instruction (conditioning input), $x_t$ the current corrupted sequence at step $t$, and $p_\theta(x_0 \mid c, x_t)$ the denoiser's token distribution over the clean sequence $x_0$ conditioned on $(c,x_t)$.
To improve the generation quality and adherence to the instruction $c$, we employ CFG~\cite{cfg}.
Let $p_\theta(x_0 \mid c, x_t)$ denote the categorical distribution over $x_0$ predicted by the neural denoiser $\hat{x}_\theta$ conditioned on $c$.
While originally proposed for continuous diffusion, we adopt the formulation adapted for discrete masked diffusion by \cite{muse}.
During training, the condition $c$ is randomly replaced with a learnable null token $\emptyset$ or an empty sequence with a fixed probability, enabling a single model to learn both conditional $p_\theta(x_0|c, x_t)$ and unconditional $p_\theta(x_0 \mid x_t)$ distributions.
During inference, the guided distribution $\tilde{p}_\theta$ is obtained by extrapolating between these two estimates:
\begin{equation}
    \tilde{p}_\theta(x_0 \mid c, x_t) \propto
    \frac{p_\theta(x_0 \mid c, x_t)^{1+w}}{
    p_\theta(x_0 \mid  x_t)^{w}},
    \label{eq:cfg_prob}
\end{equation}
where $w \ge 0$ denotes the guidance scale.
We apply this guidance in the log-space to adjust the logits before sampling:
\begin{equation}
\begin{aligned}
\log \tilde{p}_\theta(x_0 \mid c, x_t)
&= (1+w)\log p_\theta(x_0 \mid c, x_t)\\
&\quad -w\log p_\theta(x_0 \mid \emptyset, x_t)  + \text{const.}
\end{aligned}
\label{eq:cfg_logspace}
\end{equation}
A higher $w$ emphasizes the condition $c$, reducing diversity but improving relevance to the prompt.

\section{DCGC}
In this section, we introduce \textbf{DCGC}, a draft-conditioned global correction framework based on MDMs. Given a problem and an imperfect draft, DCGC generates a corrected reasoning trace through iterative denoising. The framework consists of two components. First, we use mixed-format supervised fine-tuning to train the model on both problem-only solving and draft-conditioned correction. Second, we introduce \textbf{Dynamic Dual-CFG}, an inference-time guidance mechanism that uses the problem statement and the draft as separate conditioning sources.
Dynamic Dual-CFG combines a problem-anchored guidance term with a draft-conditioned guidance term, and modulates their strength using token-level relative confidence.
% In this section, we introduce \textbf{DCGC} for solution refinement using MDMs. Our approach integrates two key components to enhance the model's ability to correct erroneous reasoning. First, we employ a unified SFT strategy that simultaneously trains the model on sandard problem-solving and refinement tasks. Second, we introduce \textbf{Dynamic Dual-CFG}, an inference-time mechanism that leverages the problem statement and the flawed solution as distinct conditioning sources. Unlike standard guidance methods, this mechanism decomposes the guidance into two hierarchical terms and dynamically modulates their strength via a confidence-aware scaling strategy.
% First, we use dual-capability supervised fine-tuning to train the model on both problem-only solving and draft-conditioned correction.

\subsection{Task Formulation}
Let $Q$ denote a problem statement, such as a math problem, coding prompt, or multiple-choice reasoning question.
Let $W$ denote an imperfect draft generated by an upstream solver, and let $G$ denote the target solution.
DCGC considers two conditional generation modes.
The first is problem-only generation, modeled as $P(G \mid Q)$.
The second is draft-conditioned correction, modeled as $P(G \mid Q, W)$, where the draft is used as auxiliary context for generating the target solution.
Our objective is to learn the draft-conditioned correction distribution $P(G \mid Q, W)$, where the model generates the target solution $G$ from the problem $Q$ and draft $W$.

% We focus on the task of solution refinement, where the goal is to correct an erroneous reasoning path into a valid solution. Formally, let $Q$ denote a problem statement (e.g., a math problem or a coding prompt) and $W$ denote a flawed solution generated by a model, which typically contains incorrect reasoning steps or a wrong final answer. In the following, we refer to this flawed solution $W$ as the draft. Our objective is to train a model to generate the correct solution, denoted as the gold solution $G$, conditioned on both the problem statement and the flawed solution. The target distribution can be formulated as $P(G \mid Q, W)$.

\subsection{Mixed-Format Supervised Fine-Tuning}
We formulate the acquisition of dual capabilities as a unified supervised learning problem~\cite{instructpix2pix}. 
We first construct a consolidated dataset by interleaving two distinct data formats, and then fine-tune the MDM on this mixture using a standard masked denoising objective. 
The training data consists of \textit{problem-only solving} pairs $(Q, G)$, where the model learns to generate the target solution $G$ directly from the problem $Q$, and \textit{draft-conditioned correction} triples $(Q, W, G)$, where the target solution is generated conditioned on both $Q$ and an imperfect draft $W$.
The problem-only pairs strengthen generation from $Q$, while the draft-conditioned triples expose the model to $W$ as auxiliary context.

% We formulate the acquisition of dual capabilities as a unified supervised learning problem~\cite{instructpix2pix}. We first construct a consolidated dataset by interleaving two distinct data formats, and then fine-tune the MDM on this mixed mixture using a standard objective. The training data consists of \textit{standard solving} pairs $(G \mid Q)$, where the model learns to generate the gold solution $G$ directly from the question $Q$, and \textit{refinement triplets} $(G \mid Q, W)$, where the target $G$ is generated conditioned on both $Q$ and a draft $W$.

\subsection{Dynamic Dual-CFG}
\label{sec:methods}
Standard classifier-free guidance uses a single conditional source.
In draft-conditioned correction, however, the problem and the draft play different roles.
The problem statement $Q$ specifies the task and answer constraints.
The draft $W$ provides an additional reasoning context that may or may not be useful for the current denoising step.
Using a single concatenated condition makes it difficult to control these two sources separately.

Dynamic Dual-CFG separates the two sources into a problem context and a joint problem-draft context.
We define the problem context as $c_{\mathrm{prob}} = Q$.
We define the joint context as $c_{\mathrm{joint}} = \{Q, W\}$.
At each denoising step, the same MDM is evaluated under three conditioning states including the unconditional context, the problem context, and the joint context.
The resulting logits are then combined through a dual guidance rule.

% Standard classifier-free guidance typically modulates generation using a single conditional source. However, solution refinement requires balancing two distinct objectives of preserving faithfulness of the original problem ($Q$)  and selectively extracting recoverable logic from the draft ($W$). Collapsing these sources into a single CFG condition is suboptimal because the two signals steer generation in different directions. Merging them blurs this control and can misguide the model away from both faithfulness to $Q$ and recoverable logic in $W$.
% % due to their diverging reliability.

% To address this, we introduce \textbf{Dynamic Dual-CFG}, which separates the guidance process into two hierarchical terms, with \textit{problem guidance} ensuring adherence to the question and \textit{refinement guidance} extracting valid patterns from the draft. We define two primary conditioning states as the problem context $c_{prob} = Q$ containing only the problem statement and the joint context $c_{joint} = \{Q, W\}$ which augments the problem with the draft. 

\paragraph{Dual-Guide Decomposition.}
We operate on the predicted clean-token logits $s_{\theta}(x_0 \mid c, x_t)$.
Let $s_{\emptyset}$, $s_{\mathrm{prob}}$, and $s_{\mathrm{joint}}$ denote the logits under the unconditional context, the problem context, and the joint context, respectively.
The final guided logits $\tilde{s}_{\theta}$ are computed as follows.

\begin{equation}
\label{eq:final_update}
\begin{aligned}
    \tilde{s}_{\theta}
    =
    &s_{\mathrm{prob}} + \underbrace{S_1 \odot (s_{\mathrm{prob}} - s_{\emptyset})}_{\text{Problem-Anchored Guidance}}\\
    & +{s_{\mathrm{joint}}} + \underbrace{S_2 \odot (s_{\mathrm{joint}} - s_{\mathrm{prob}})}_{\text{Relative Draft Residual}} .
\end{aligned}
\end{equation}
Here, $S_1$ and $S_2$ are position-wise scaling factors, and $\odot$ denotes element-wise scaling over sequence positions with broadcasting over the vocabulary dimension.
The first term applies standard CFG around the problem context and anchors the generation to $Q$.
The remaining terms integrate the draft-conditioned information using the joint prediction $s_{\mathrm{joint}}$ alongside a relative residual $S_2 \odot (s_{\mathrm{joint}} - s_{\mathrm{prob}})$. This residual formulation captures how the draft context shifts the model's prediction trajectory compared to the problem-only branch. Overall, this decomposition allows DCGC to robustly combine problem anchoring with draft-guided correction, where $S_2$ dictates the amplification of the draft's influence.
The detailed derivation is provided in Appendix~\ref{app:derivation}.

\paragraph{Confidence-Modulated Scaling.}
Fixed guidance scales use the same strength across all positions and examples~\cite{llada,mdm}.
DCGC instead computes token-level guidance scales from the model's confidence under the problem and joint contexts.
We define confidence as the maximum probability in the predicted token distribution.
For each token $i$, we compute the confidence scores as follows.

\begin{equation}
\label{eq:confidence}
\begin{split}
    C_{\mathrm{prob}}^{(i)}
    &= \max_v \; \mathrm{Softmax}(s_{\mathrm{prob}}^{(i)})_v, \\
    C_{\mathrm{joint}}^{(i)}
    &= \max_v \; \mathrm{Softmax}(s_{\mathrm{joint}}^{(i)})_v .
\end{split}
\end{equation}

Based on these scores, we define the position-wise scaling factors.

\begin{equation}
\label{eq:scaling}
\begin{split}
    S_1^{(i)} &= \alpha \cdot C_{\mathrm{prob}}^{(i)}, \\
    S_2^{(i)} &= \beta \cdot \mathrm{ReLU}
    \left(
    C_{\mathrm{joint}}^{(i)} - C_{\mathrm{prob}}^{(i)}
    \right),
\end{split}
\end{equation}
where $\alpha$ and $\beta$ are scalar hyperparameters.
The problem guidance scale $S_1$ increases with problem-context confidence, which strengthens the problem-anchored term at positions where the model is confident under $Q$.
The draft residual scale $S_2$ depends on the relative confidence gap between the joint and the problem-only branch.
When the joint branch has higher confidence than the problem-only branch at position $i$, DCGC amplifies the residual direction $s_{\mathrm{joint}}^{(i)} - s_{\mathrm{prob}}^{(i)}$.
When the joint branch does not provide a confidence gain, the additional residual term at that position is set to zero.

This design uses relative confidence as an internal signal for modulating guidance strength.
The joint-context prediction remains part of the combined logits through $s_{\mathrm{joint}}$, while the residual amplification is activated only when the joint branch provides higher confidence than the problem-only branch.
The resulting scaling factors are applied independently across sequence positions in Eq.~\ref{eq:final_update}.

\section{Experimental Setups}

We describe the implementation of DCGC and the baselines used for comparison. Dataset details, inference and training hyperparameters, and prompt templates are deferred to Appendix~\ref{app:exp_details} and Appendix~\ref{app:prompts}, respectively.

\paragraph{SFT Dataset Curation.}
To equip the model with both problem-only and draft-conditioned generation capabilities, we construct a composite SFT dataset across mathematics, coding, and general reasoning domains.
The training mixture contains two types of examples.
The first type consists of standard solving pairs $(Q, G)$, where the model generates the gold solution from the problem alone.
The second type consists of draft-conditioned correction triples $(Q, W, G)$, where the model generates the gold solution using both the problem and an imperfect draft.

Since imperfect drafts $W$ are not readily available for all domains, we use domain-specific curation strategies.
For mathematics, we utilize labeled incorrect trajectories derived from the Math-Sheperd dataset~\cite{math-shepherd}.
For coding and general reasoning, we collect model-generated outputs or failure cases from open-source models and use them as imperfect drafts.
To improve training stability, we filter examples whose full prompt exceeds 1,028 tokens.
The final filtered dataset was split into training and validation sets, with 5\% of the filtered examples reserved for validation.

\paragraph{Benchmarks and Evaluation Protocol.}
We evaluate our method on diverse benchmarks, including GSM8K~\cite{gsm8k} and MATH-500~\cite{math} for mathematics, MBPP-test~\cite{mbpp} and HumanEval~\cite{humev} for code generation, and MMLU-STEM and MMLU-Pro~\cite{mmlu} for general reasoning. 
Our main experiments use solver-failure hard sets.
For each benchmark, we run Llama-3.1-8B-Instruct on the test set and retain only the instances where the solver produces an incorrect answer.
The generated incorrect answer is used as the draft $W$.
This protocol evaluates conditional correction performance on failed reasoning attempts.
% The resulting hard sets contain 216 examples for GSM8K, 282 for MATH, 216 for MBPP, 69 for HumanEval, 956 for MMLU-STEM, and 6,675 for MMLU-Pro.

We also evaluate DCGC on full test sets in Section~\ref{sec:realscen}, where ground-truth labels are not used to determine whether refinement is needed.
This full-set evaluation measures the practical effect of applying DCGC under a gold-agnostic selective refinment strategy.

\begin{table*}[t]
    \centering
    \caption{\textbf{Overall Performance.} Performance comparison across reasoning benchmarks. DCGC combines mixed-format SFT with Dynamic Dual-CFG and achieves the highest average accuracy, with the best score on five of the
    six benchmarks. \textbf{Bold} and \underline{underlined} indicate the best and second-best scores, respectively.}
    \label{tab:main_results1}
    \resizebox{\textwidth}{!}{ % 표가 너무 크면 0.9\textwidth 등으로 조절
    \begin{tabular}{l l c c c c c c c}
        \toprule
        \multirow{2}{*}{\textbf{Method}} & \multirow{2}{*}{\textbf{Model}} & \multicolumn{2}{c}{\textbf{Math}} & \multicolumn{2}{c}{\textbf{Code}} & \multicolumn{2}{c}{\textbf{Knowledge \& Reasoning}} &\multirow{2}{*}{\textbf{Avg.}} \\
        \cmidrule(lr){3-4} \cmidrule(lr){5-6} \cmidrule(lr){7-8}
         & & \textbf{GSM8K} & \textbf{MATH} & \textbf{MBPP} & \textbf{HumanEval} & \textbf{MMLU-STEM} & \textbf{MMLU-Pro} \\
        \midrule
        \multicolumn{9}{l}{\textit{\textbf{LLM Baselines}}} \\
        Self-Refine   & LLaMA  & 17.1 & 0.4  & \textbf{11.1} & \underline{11.6} & 11.3 & 5.8 & 10.1\\
        Self-Refine   & $\text{LLaMA}_{\text{SFT}}$  &26.4 	&11.0 & 8.3 & 10.1 & 26.3 & 15.9 	& 16.3 \\
        Self-Refine  & Mistral & 5.6 & 0.7  & 1.4 & 8.7 & 4.2 & 3.0 &3.9\\
        \midrule
        \multicolumn{9}{l}{\textit{\textbf{MDM variants}}} \\
        Standard Sampling & LLaDA  & 7.4 & 9.9 & 0.0 & 0.0 & 5.7 & 5.1 &4.7 \\
        Dynamic Dual-CFG  & LLaDA & 16.7 & 9.9 & 3.2 & 4.4 & 19.7 & 13.4  &11.2\\
        Standard Sampling & $\text{LLaDA}_{\text{SFT}}$ & \underline{32.4} & \underline{17.4} & 5.1 & 8.7 & \underline{30.9} & \underline{19.2} & \underline{18.9}\\
        \midrule
        \multicolumn{9}{l}{\textit{\textbf{Ours}}} \\
        \textbf{DCGC} & \textbf{$\text{LLaDA}_{\text{SFT}}$} & \textbf{44.9} & \textbf{22.3} & \underline{10.7} & \textbf{13.1} & \textbf{35.7} & \textbf{22.5} & \textbf{24.8}\\
        \bottomrule
    \end{tabular}
    }
\end{table*}

\paragraph{Baselines and variants.}
Our main comparison evaluates DCGC against controlled MDM variants built on the same LLaDA-8B-Instruct backbone~\cite{llada}. We also include tool-free autoregressive self-refinement baselines as references to prior self-correction methods. For these baselines, we use Llama-3.1-8B-Instruct~\cite{llama3}, its SFT-adapted variant trained on the same dataset, and Mistral-7B-v1~\cite{mistral}, each evaluated with the Self-Refine pipeline~\cite{selfrefine}.

% For MDM variants, we consider both the base LLaDA backbone and its dual-capability SFT variant, denoted as $\text{LLaDA}_{\text{SFT}}$.
For MDM variants, we consider the base LLaDA backbone and its SFT-adapted variant, $\text{LLaDA}_{\text{SFT}}$, trained on our mixed-format SFT dataset.
We first include \textit{Standard Sampling}, which performs masked diffusion decoding without classifier-free guidance.
We evaluate this setting for both the base LLaDA backbone and $\text{LLaDA}_{\text{SFT}}$ to measure the effect of dual-capability SFT.
We also apply \textit{Dynamic Dual-CFG} to the base backbone to examine how much the guidance mechanism contributes before task-specific adaptation.

For the SFT-adapted backbone, we compare DCGC with several guidance variants.
\textit{Single-CFG (Problem)} applies CFG using only the problem statement $Q$.
\textit{Single-CFG (Joint)} applies CFG using the joint problem-draft context $(Q,W)$.
\textit{Dual-CFG (Static)} separates the problem-only and problem-draft branches but uses constant scaling.
\textit{Dual-CFG (Independent)} uses a linear scaling rule based on the confidence of the joint branch.
Finally, \textit{DCGC (Relative)} uses the proposed relative ReLU scaling based on the confidence gap between the joint branch and the problem-only branch.

\paragraph{Implementation Details.}
We implement DCGC using the LLaDA-8B-Instruct backbone. To ensure parameter efficiency, we employ low-rank adapation (LoRA)~\cite{lora} for fine-tuning. The model is trained for 3 epochs on 2 NVIDIA A100 GPUs. For inference, we set the generation length to 256 tokens (512 for MMLU). All MDM variants are evaluated under the same hard-set protocol and use the same block diffusion setting with block size 32~\cite{block-diffusion}.
Crucially, the guidance hyperparameters for our dynamic dual-CFG mechanism are strictly fixed to $\alpha = 0.5$ and $\beta = 1.0$, calibrated solely on the GSM8K validation split and held constant across all benchmarks. In contrast, for the constant-scaling baselines, we sweep the guidance weights from $0.5$ to $2.0$ with a step size of $0.5$ and report the maximum score achieved. This evaluation protocol intentionally frames the baseline performance at its empirical upper limit, ensuring a highly conservative assessment of our proposed method. 
Additional training and inference details are provided in Appendix~\ref{app:exp_details}.

\section{Results}
\subsection{Main Results}
\label{sec:main_results}

\paragraph{Overall performance.}
Table~\ref{tab:main_results1} and Table~\ref{tab:main_results2} summarize the overall results, showing that DCGC achieves the strongest overall performance across the evaluated benchmarks.
The largest gains appear in mathematical reasoning.
On GSM8K, DCGC attains 44.9\%, compared to 26.4\% from the strongest autoregressive \textit{Self-Refine} baseline and 43.9\% from the strongest non-DCGC LLaDA variant. 
Similarly on MATH, DCGC reaches 22.3\%, surpassing the strongest autoregressive baseline at 11.0\% and the best non-DCGC LLaDA variant at 18.4\%.

This advantage extends to code and knowledge-heavy tasks, where DCGC obtains 13.1\% on HumanEval and achieves top scores of 35.7\% on MMLU-STEM and 22.5\% on MMLU-Pro.
While MBPP performance is comparable to the strongest baseline (10.7\% vs. 11.1\%), we attribute the larger gains on HumanEval to its richer problem specification.
HumanEval includes function signatures and docstring-style instructions that provide more explicit constraints for correction than MBPP's brief natural-language prompts.

Overall, these results indicate that DCGC is an effective draft-conditioned global correction module when combined with dual-capability SFT and relative dual guidance.
We provide a representative step-by-step example in Table~\ref{app-tab:qualitative_example}, and present qualitative comparisons between baselines and DCGC in Appendix~\ref{app-sec:qual}.

\paragraph{Effect of SFT.}

We first examine the role of SFT by comparing \textit{standard sampling} using the base LLaDA-8B-Instruct model to the SFT-adapted variant.
Without SFT, the pre-trained diffusion model performs poorly, achieving only 4.7 average accuracy and 7.4\% on GSM8K.
In contrast, SFT substantially improves performance across tasks.
With standard sampling, $\text{LLaDA}_{\text{SFT}}$ reaches 18.9 average accuracy, including 32.4\% on GSM8K, 30.9\% on MMLU-STEM, and 19.2\% on MMLU-Pro.
These results support the view that dual-capability SFT equips the model with both problem-only solving ability and draft-conditioned generation ability.
This training stage is therefore an important foundation for DCGC, since the model must remain grounded in the problem statement while still being able to use draft-derived cues when they are informative.

\begin{table*}[t]
    \centering
    \caption{\textbf{Impact of guidance scaling strategies.} DCGC (Relative) utilizes a dynamic ReLU-based scaling to selectively leverage dual conditions, consistently outperforming static and independent variants across all benchmarks. \textbf{Bold} and \underline{underlined} values indicate the best and second-best scores, respectively.}
    \label{tab:main_results2}
    \resizebox{\textwidth}{!}{
    \begin{tabular}{l c l l c c c c c c c}
        \toprule
        \multirow{2}{*}{\textbf{Method}} & \multicolumn{2}{c}{\textbf{Guidance Branch}} & \multirow{2}{*}{\textbf{Scaling}} & \multicolumn{6}{c}{\textbf{Scores}} & \multirow{2}{*}{\textbf{Avg.}}\\
        \cmidrule(lr){2-3} \cmidrule(lr){5-10}
         & Q & (Q, W) & & \textbf{GSM8K} & \textbf{MATH} & \textbf{MBPP} & \textbf{HEval} & \textbf{STEM} & \textbf{Pro} \\
        \midrule
        Standard Sampling & - & - & - & 32.4 & 17.4 & 5.1 & 8.7 & 30.9 & 19.2 & 18.9\\
        \midrule
        \multicolumn{11}{l}{\textit{Single-Condition}} \\
        Single-CFG (\textit{Problem}) & \checkmark & - & Constant & \underline{43.9} & 17.4  & 6.9 & \underline{10.1} & \underline{34.9} & 20.8 & \underline{22.4} \\
        Single-CFG (\textit{Joint}) & - & \checkmark & Constant & 36.6 & 15.6 & 5.6 & 8.7 & 32.9 & \underline{21.4}  & 20.1\\
        \midrule
        \multicolumn{11}{l}{\textit{Dual-Condition}} \\
        Dual-CFG (\textit{Static})      & \checkmark & \checkmark & Constant & 37.9 & 15.9 & 5.6 & 7.3 & 33.5 & 20.9 & 20.2\\
        Dual-CFG (\textit{Independent}) & \checkmark & \checkmark & Linear   & 38.4 & \underline{18.4} & \underline{9.3} & 8.7 & 33.7 & 20.6 &21.5 \\
        \textbf{DCGC (\textit{Relative})} & \checkmark & \checkmark & \textbf{ReLU} & \textbf{44.9} & \textbf{22.3} & \textbf{10.7} & \textbf{13.1} & \textbf{35.7} & \textbf{22.5} &\textbf{24.8} \\
        \bottomrule
    \end{tabular}
    }
\end{table*}

\paragraph{Limits of Single-Condition Guidance.}
Beyond training, we examine the role of Dynamic Dual-CFG by comparing standard sampling against single-condition CFG and dual-branch CFG variants. This allows us to separate the effect of inference-time guidance from the effect of conditioning design. 
Applying Dynamic Dual-CFG to the base LLaDA model improves the average accuracy from 4.7 to 11.2, with GSM8K increasing from 7.4\% to 16.7\%.
This shows that inference-time guidance can provide a useful correction signal even before task-specific adaptation.

On the SFT-adapted backbone, \textit{Single-CFG (Problem)} improves over standard sampling from 18.9 to 22.4, showing that strong problem conditioning can recover many failed cases.
This problem-only branch provides a useful problem-conditioned reference for generation.
However, problem-only guidance is not sufficient to obtain the best performance.
DCGC improves over \textit{Single-CFG (Problem)} on all six benchmarks, with especially clear gains on MATH, MBPP, and HumanEval.
This shows that draft-conditioned information can complement problem-only guidance when its residual contribution is properly controlled.

This benefit does not come from simply adding the draft.
Although the joint condition contains more information, \textit{Single-CFG (Joint)}, which uses the concatenated problem-draft context, drops to 20.1 average accuracy.
This indicates that imperfect drafts can introduce distracting context when their contribution is not separated from the problem statement.
Therefore, effective use of the draft requires separating the problem-conditioned and draft-conditioned signals, which motivates the dual-guide design of DCGC.

\paragraph{Effect of relative scaling.}
As illustrated in Table~\ref{tab:main_results2}, after separating the problem and joint problem-draft branches, we evaluate how the draft-conditioned residual should be scaled.
\textit{Dual-CFG (Static)} uses constant scaling for the separated branches, but reaches only 20.2 average accuracy, remaining below the problem-only baseline at 22.4.
This shows that uniformly enforcing the draft's influence across the entire problem is insufficient.

\textit{Dual-CFG (Independent)} improves over static dual guidance on nearly all benchmarks, reaching 21.5 average accuracy, but it still underperforms DCGC.
This variant scales the draft-conditioned branch based on the confidence of the joint context alone.
However, joint confidence is not always a reliable signal for amplifying draft-conditioned information, since an imperfect draft can still make the joint branch confident.

In contrast, DCGC compares the joint problem-draft branch against the problem-only branch and uses a relative ReLU confidence gap to scale the residual direction between them.
When the joint branch provides a confidence gain over the problem-only branch, DCGC increases the draft-conditioned residual.
When it provides little additional support, the extra residual amplification is reduced.
As a result, DCGC achieves 24.8 average accuracy and obtains the best scores among LLaDA variants across all benchmarks.
These results show that relative scaling is important for balancing problem conditioning with draft-conditioned residual guidance.
We report additional results with alternative scaling functions in Appendix~\ref{app:activation}.

\subsection{Draft Relevance and Selective Reuse}
\label{sec:draft_perturbation}

While our main results demonstrate that DCGC effectively improves reasoning accuracy, it is important to verify whether the model genuinely utilizes the provided draft or merely ignores it and re-solves the problem from scratch. To address this, Table~\ref{tab:draft_perturbation} evaluates whether DCGC is sensitive to the relevance of the draft condition.
To isolate the effect of the draft, we keep the problem fixed and replace only the draft input.
The \textit{Original} setting uses the draft generated for the same problem, while \textit{Shuffled} and \textit{Domain-shifted} replace it with irrelevant drafts from the same task or a different task.
Across all evaluated tasks, the original draft achieves the highest accuracy and draft reuse ratio (FRR). For example, on MATH, accuracy drops from 22.3 with the original draft to 10.3 and 11.0 under shuffled and domain-shifted drafts, respectively. Similar patterns are observed on other benchmarks, with particularly large changes on the MBPP code generation task. 

Overall, these perturbation results show that DCGC does not treat all drafts equally.
Instead, it reuses aligned and informative draft content while suppressing reuse when the draft becomes irrelevant or mismatched.
% Relevant drafts lead to higher correction accuracy and greater draft reuse, while irrelevant drafts lead to lower reuse and more minimal-reuse outputs.
Together with Table~\ref{tab:main_results2}, this supports the role of relative dual guidance in using draft-derived information beyond problem-only guidance.

\begin{table}[t]
    \centering
    \resizebox{\columnwidth}{!}{
    \scriptsize
    \setlength{\tabcolsep}{4pt}
    \begin{tabular}{l l c c}
    \toprule
    Task & Draft Condition & Acc. $\uparrow$ &  FRR $\downarrow$ \\
    \midrule
    GSM8K 
        & Original       & \textbf{44.9}  & \textbf{27.3} \\
        & Shuffled       & 38.0 & 90.3 \\
        & Domain-shifted & 40.7 & 98.6 \\
    \midrule
    MATH 
        & Original       & \textbf{22.3} & \textbf{66.0} \\
        & Shuffled       & 10.3  & 96.8 \\
        & Domain-shifted & 11.0 & 96.8 \\
    \midrule
    MBPP 
        & Original       & \textbf{10.7}  & \textbf{3.2} \\
        & Shuffled       & 10.6  & 17.6 \\
        & Domain-shifted & 7.4   & 100.0 \\
    \midrule
    MMLU-STEM 
        & Original       & \textbf{35.7}  & \textbf{36.5} \\
        & Shuffled       & 32.1   & 89.0 \\
        & Domain-shifted & 30.0  & 88.8 \\
    \bottomrule
    \end{tabular}}
    \caption{
    \textbf{Draft relevance perturbation.}
    We keep the problem fixed and perturb only the draft condition.
    \textit{Original} uses the draft generated for the same problem, 
    \textit{Shuffled} uses an unrelated draft from the same task, and 
    \textit{Domain-shifted} uses a draft from a different task.
    FRR (Full Regeneration Rate) denotes the fraction of samples where the generated output has low token overlap with the draft, retaining fewer original draft tokens.
    % FRR (Full Regeneration Rate) denotes the fraction of samples where the model effectively ignores the draft by retaining fewer original tokens than a threshold. 
    % A lower FRR on the \textit{Original} draft indicates successful utilization of relevant context.
    }
    \label{tab:draft_perturbation}
\end{table}

% These results indicate that DCGC does not blindly copy from the draft. Instead, it reuses aligned and informative draft content while suppressing reuse when the draft becomes irrelevant or mismatched.
% On GSM8K, Draft Reuse drops from 15.7 to 5.9 and 2.9 under the two perturbations, while Minimal Reuse increases from 27.3 to 90.3 and 98.6.
% This pattern indicates that DCGC changes its behavior substantially when the draft is no longer aligned with the problem.
% The effect is especially clear in structured code generation.
% On MBPP, Draft Reuse decreases from 57.5 with the original draft to 19.0 under shuffled drafts and 1.0 under domain-shifted drafts.
% This suggests that relevant code drafts can provide reusable structure such as function format, boilerplate, or implementation patterns.
% When the draft is irrelevant, this reuse largely disappears.
\begin{table}[t]
\centering
\resizebox{\columnwidth}{!}{%
\begin{tabular}{l|cc|cc}
\toprule
\multirow{2}{*}{\textbf{Method}} 
& \multicolumn{2}{c|}{\textbf{Full Set}} 
& \multicolumn{2}{c}{\textbf{Correction Set}} \\
& \textbf{Acc.} & \textbf{$\Delta$} 
& \textbf{Acc.} & \textbf{$\Delta$} \\
\midrule
\multicolumn{5}{c}{\textbf{MATH} \textit{(Correction Ratio $\sim$64\%, Maj$\geq$3@5 21.00)}} \\
\midrule
Llama-3-8B (Maj@5) & 41.19 & - & 20.4 & - \\
+ Self-Refine & 41.20 & \textcolor{blue}{(+0.01)} & 20.4 & \textcolor{blue}{(+0.0)} \\
+ LLaDA (Standard) & 40.40 & \textcolor{red}{(-0.79)} & 19.1 & \textcolor{red}{(-1.3)} \\
+ $\text{LLaDA}_{\text{SFT}}$ (Standard) & 42.60 & \textcolor{blue}{(+1.41)} & 22.6 & \textcolor{blue}{(+2.2)} \\
+ $\text{LLaDA}_{\text{SFT}}$ (Single-CFG Q) & 40.80 & \textcolor{red}{(-0.39)} & 19.7 & \textcolor{red}{(-0.6)} \\
\textbf{+ DCGC} & \textbf{43.20} & \textcolor{teal}{\textbf{(+2.01)}} & \textbf{23.5} & \textcolor{teal}{\textbf{(+3.1)}} \\

\midrule
\multicolumn{5}{c}{\textbf{GSM8K} \textit{(Correction Ratio $\sim$9\%, Maj$\geq$3@5 78.47)}} \\
\midrule
Llama-3-8B (Maj@5) & 86.80 & - & 38.1 & - \\
+ Self-Refine & 85.97 & \textcolor{red}{(-0.83)} & 28.8 & \textcolor{red}{(-9.3)} \\
+ LLaDA (Standard) & 85.37 & \textcolor{red}{(-1.43)} & 22.0 & \textcolor{red}{(-16.1)} \\
+ $\text{LLaDA}_{\text{SFT}}$ (Standard) & 85.75 & \textcolor{red}{(-1.05)} & 26.3 & \textcolor{red}{(-11.9)} \\
+ $\text{LLaDA}_{\text{SFT}}$ (Single-CFG Q) & 86.20 & \textcolor{red}{(-0.60)} & 31.4 & \textcolor{red}{(-6.8)} \\
\textbf{+ DCGC} & \textbf{86.96} & \textcolor{teal}{\textbf{(+0.16)}} & \textbf{39.8} & \textcolor{teal}{\textbf{(+1.7)}} \\
\midrule
\multicolumn{5}{c}{\textbf{MMLU-STEM} \textit{(Correction Ratio $\sim$7\%, Maj$\geq$3@5 64.13)}} \\
\midrule
Llama-3-8B (Maj@5) & 68.66 & - & 26.4 & - \\
+ Self-Refine & 68.79 & \textcolor{blue}{(+0.13)} & 28.2 & \textcolor{blue}{(+1.9)} \\
+ LLaDA (Standard) & 68.47 & \textcolor{red}{(-0.19)} & 23.6 & \textcolor{red}{(-2.8)} \\
+ $\text{LLaDA}_{\text{SFT}}$ (Standard) & 68.89 & \textcolor{blue}{(+0.23)} & 29.6 & \textcolor{blue}{(+3.2)} \\
+ $\text{LLaDA}_{\text{SFT}}$ (Single-CFG Q) & 68.82 & \textcolor{blue}{(+0.16)} & 28.7 & \textcolor{blue}{(+2.3)} \\
\textbf{+ DCGC} & \textbf{69.17} & \textcolor{teal}{\textbf{(+0.51)}} & \textbf{33.8} & \textcolor{teal}{\textbf{(+7.4)}} \\
\bottomrule
\end{tabular}%
}
\caption{
\textbf{Gold-agnostic selective correction on full test sets.}
Only low-consensus examples are selected for correction. Headers indicate the correction ratio and high-consensus (Maj$\geq$3@5) accuracy. Parentheses denote $\Delta$ relative to the no-correction Maj@5 baseline.}

% Low-consensus examples without a $\geq$3 majority among five samples are routed to correction, while the remaining examples keep the upstream Maj@5 answer.
% Dataset headers report the routed ratio and the accuracy of the high-consensus Maj$\geq$3@5 subset.
% Values in parentheses denote $\Delta$ relative to the no-correction Maj@5 pipeline.
\label{tab:real_world}
\end{table}

\begin{table}[t]
\centering
\small
\setlength{\tabcolsep}{5pt}
\begin{tabular}{lcc}
\toprule
Method & Easy & Hard \\
\midrule

\multicolumn{3}{c}{\textbf{GSM8K} (Easy: 45, Hard: 73)} \\
\midrule
DCGC                         & \textbf{53.33} & \textbf{31.51} \\
LLaDA Standard                & 42.22          & 9.59           \\
LLaDA$_{\mathrm{SFT}}$ Standard
                              & 44.44          & 15.07          \\
LLaDA$_{\mathrm{SFT}}$ cond $Q$
                              & 40.00          & 26.03          \\
LLaMA-8B-Instruct             & 46.67          & 17.81          \\

\midrule
\multicolumn{3}{c}{\textbf{MATH} (Easy: 65, Hard: 254)} \\
\midrule
DCGC                         & 44.62          & \textbf{18.11} \\
LLaDA Standard                & 43.08          & 12.99          \\
LLaDA$_{\mathrm{SFT}}$ Standard
                              & 47.69          & 16.14          \\
LLaDA$_{\mathrm{SFT}}$ cond $Q$
                              & 41.54          & 14.17          \\
LLaMA-8B-Instruct             & \textbf{50.77} & 12.20          \\

\midrule
\multicolumn{3}{c}{\textbf{MMLU-STEM} (Easy: 57, Hard: 159)} \\
\midrule
DCGC                         & 40.35          & \textbf{31.45} \\
LLaDA Standard                & \textbf{49.12} & 14.47          \\
LLaDA$_{\mathrm{SFT}}$ Standard
                              & 40.35          & 25.79          \\
LLaDA$_{\mathrm{SFT}}$ cond $Q$
                              & 31.58          & 27.67          \\
LLaMA-8B-Instruct             & 42.11          & 23.27          \\
\bottomrule
\end{tabular}
\caption{\textbf{Post-hoc breakdown of accuracy on the low-consensus correction
subset.} Easy and hard denote instances for which the initial upstream
solution is correct and incorrect, respectively.}
\label{tab:easy-hard-breakdown}
\end{table}

% \begin{table}[t]
% \centering
% \small
% \resizebox{\columnwidth}{!}{%
% \begin{tabular}{lcc cc cc}
% \toprule
% & \multicolumn{2}{c}{GSM8K}
% & \multicolumn{2}{c}{MATH}
% & \multicolumn{2}{c}{MMLU-STEM} \\
% \cmidrule(lr){2-3} \cmidrule(lr){4-5} \cmidrule(lr){6-7}
% Method
% & Easy (45) & Hard (73)
% & Easy (65) & Hard (254)
% & Easy (57) & Hard (159) \\
% \midrule
% DCGC
% & \textbf{53.33} & \textbf{31.51}
% & 44.62 & \textbf{18.11}
% & 40.35 & \textbf{31.45} \\
% LLaDA Standard
% & 42.22 & 9.59
% & 43.08 & 12.99
% & \textbf{49.12} & 14.47 \\
% LLaDA$_{\mathrm{SFT}}$ Standard
% & 44.44 & 15.07
% & 47.69 & 16.14
% & 40.35 & 25.79 \\
% LLaDA$_{\mathrm{SFT}}$ cond $Q$
% & 40.00 & 26.03
% & 41.54 & 14.17
% & 31.58 & 27.67 \\
% LLaMA-8B-Instruct
% & 46.67 & 17.81
% & \textbf{50.77} & 12.20
% & 42.11 & 23.27 \\
% \bottomrule
% \end{tabular}%
% }
% \caption{Post-hoc breakdown of accuracy on the low-consensus correction
% subset. Easy and hard denote instances for which the initial upstream
% solution is correct and incorrect, respectively. }
% \label{tab:easy-hard-breakdown}
% \end{table}

\subsection{Evaluation in Gold-Agnostic Settings}
\label{sec:realscen}
We further evaluate DCGC in a gold-agnostic setting where ground-truth answers are unavailable at test time. We use Llama-3.1-8B with self-consistency~\cite{selfconsis} over five samples as the upstream generator.
If no answer receives at least three votes among the five samples, the query is marked as low-consensus and sent to a correction module.
Otherwise, the self-consistency answer is accepted.
For these correction cases, we use the first generated reasoning path as the draft $W$.
% To simulate this, we construct a collaborative pipeline using a Llama-3-8B generator with \textit{Self-Consistency}~\cite{selfconsis}. Specifically, we sample five reasoning paths and flag queries with low consensus ($Maj@5 < 3/5$) as \textit{uncertain}, while reliable queries are directly accepted. For the refinement stage, we utilize the first generated path among these five samples as the draft $W$.

To rigorously evaluate the benefit of correction, we employ Majority Voting ($Maj@5$) as a strong baseline that inherently aggregates the model's best possible predictions.
We additionally report $Maj{\geq}3@5$, which counts a prediction as correct only when the selected answer appears at least three times and matches the gold answer. 
This diagnostic characterizes the strength of the upstream consensus.
Surpassing $Maj@5$ implies correcting instances where the generator's internal consensus is either scattered or fundamentally incorrect. This is a task significantly harder than improving upon a single sample ($pass@1$).
% Here, $Maj@5$ selects the most frequent answer among five sampled solutions and evaluates it against the gold answer. 

Table~\ref{tab:real_world} reports the full-set accuracy as the primary metric. Overall, DCGC consistently improves the final accuracy across all three benchmarks, GSM8K, MATH, and MMLU-STEM. Notably, the results highlight that correction under low consensus is highly unstable for conventional baselines. For instance, on GSM8K, standard methods like \textit{Self-Refine} actually degrade the correction-subset accuracy by $9.3\%$ compared to the $Maj@5$ baseline. This performance drop underscores the inherent difficulty of reliably revising uncertain drafts without ground-truth feedback. In contrast, DCGC consistently improves correction-subset accuracy over the $Maj@5$ baseline, even though the corresponding full-set gains are modest on GSM8K and MMLU-STEM due to their small correction ratios.

To further characterize performance on the low-consensus correction subset, we divide it according to whether the initial upstream solution is correct. We refer to instances with a correct initial solution as \textit{easy} and those with an incorrect initial solution as \textit{hard}; these labels are used only for post-hoc analysis and are not available to the correction methods at test time. Table~\ref{tab:easy-hard-breakdown} reports the resulting breakdown.

DCGC achieves the highest accuracy on the hard subset for all three benchmarks, reaching 31.51 on GSM8K, 18.11 on MATH, and 31.45 on MMLU-STEM. These results show that its gains in the gold-agnostic setting arise from more reliable correction of initially incorrect, low-consensus outputs. Meanwhile, DCGC remains competitive on the easy
subset (53.33, 44.62, and 40.35 on GSM8K, MATH, and MMLU-STEM, respectively), indicating that its hard-case improvements are not explained solely by an indiscriminate trade-off against initially correct cases.

% Table~\ref{tab:real_world} reports full-set accuracy as the primary metric. DCGC improves the final full-set accuracy on all three benchmarks, from 86.80 to 86.96 on GSM8K, from 41.19 to 43.20 on MATH, and from 68.66 to 69.17 on MMLU-STEM. 
% The results show that correction under low consensus can be unstable for several baselines. As shown in the GSM8K results, standard baselines such as \textit{Self-Refine} decrease routed-subset accuracy by $9.3\%$ compared to the $Maj@5$ baseline. This suggests that low-consensus cases are difficult to improve reliably, since the model must revise an uncertain draft without access to ground-truth feedback.
% In comparison, DCGC achieves positive routed-subset gains across all benchmarks and improves full-set accuracy. 
% Although the full-set gains are modest on GSM8K and MMLU-STEM due to the small routed ratios, DCGC improves routed-subset accuracy by $7.4\%$ on MMLU-STEM and $3.1\%$ on MATH compared to the $Maj@5$ baseline. 
% These results suggest that applying DCGC selectively to low-consesus cases produces a positive net effect on the full test set.

\section{Conclusion}

In this paper, we presented DCGC, a masked-diffusion framework for draft-conditioned global correction of complex reasoning traces. DCGC leverages the synergy between Supervised Fine-Tuning and Dynamic Dual-CFG, enabling the model to generate under both problem-only and joint problem-draft contexts while adaptively modulating the draft-conditioned residual signal. 
Extensive experiments across diverse benchmarks demonstrate that DCGC significantly outperforms strong baselines on reasoning tasks, and remains effective in realistic settings without oracle feedback. 
DCGC's ability to extract constructive draft information via relative dual guidance highlights the promising paradigm of combining autoregressive models for initial generation with masked diffusion models for global correction to build reliable, self-correcting reasoning systems.
% DCGC benefits from relevant draft information beyond problem-only guidance while remaining grounded in the problem statement through relative dual guidance.
% This suggest that combining autoregressive models for initial generation with masked diffusion models for global correction is a promising direction for building more reliable and self-correcting reasoning systems.
% Extensive experiments across diverse benchmarks demonstrate that DCGC significantly outperforms strong baselines on challenging tasks such as MATH and HumanEval, and remains effective in realistic settings without oracle feedback.

\section*{Limitations}
\label{sec:limitations}
Due to computational resource constraints during supervised fine-tuning, we applied a strict length filter and discarded samples exceeding 1,028 tokens. While this length is sufficient for the reasoning benchmarks evaluated in this work, such as GSM8K, MATH, and standard MMLU tasks, the current instantiation of DCGC may be limited in extremely long-context scenarios. Additionally, our gold-agnostic correction strategy uses self-consistency as a simple uncertainty signal, and more sophisticated triggering criteria may further improve efficiency and reliability.

% DCGC should not be interpreted as a truth-verifying correction module. 
% The relative confidence gap used by Dynamic Dual-CFG is an internal model signal, not a guarantee of factual or logical correctness. 
% Highly fluent but incorrect drafts may still increase joint confidence and mislead the model, especially when the problem-only branch is uncertain. 
% Thus, DCGC is best suited to non-adversarial drafts generated for the same problem under a selective routing setting, and future work may combine draft-conditioned global correction with external verifiers or execution-based checks.

\section*{Acknowledgments}
This work is in part supported by the National Research Foundation of Korea (NRF, RS-2024-00451435(20\%), RS-2024-00413957(20\%)), Institute of Information \& communications Technology Planning \& Evaluation (IITP, RS-2025-02305453(15\%), RS-2025-02273157(15\%), RS-2025-25442149(15\%) RS-2021-II211343(15\%)) grant funded by the Ministry of Science and ICT (MSIT), Institute of New Media and Communications(INMAC), and the BK21 FOUR program of the Education, Artificial Intelligence Graduate School Program (Seoul National University), and Research Program for Future ICT Pioneers, Seoul National University in 2026.

\newpage
\bibliography{custom}

\newpage
\appendix

\section{Derivation of Dual-Guide CFG}
\label{app:derivation}

In this section, we provide the mathematical derivation of the Dual-Guide Decomposition objective presented in Eq.~\ref{eq:final_update}. We show how the logit-space manipulation corresponds to composing a target distribution from multiple guidance signals.

\paragraph{Relationship between Logits and Probabilities.}
In diffusion language models, the network output $s_\theta(x_0 \mid c, x_t)$ represents the unnormalized logits. The probability distribution is obtained via the softmax function:
\begin{equation}
    p_\theta(x_0 \mid c, x_t) = \frac{\exp(s_\theta(x_0 \mid c, x_t))}{Z(c, x_t)},
\end{equation}
where $Z(c, x_t)$ is the normalization constant. In the log-space, this implies:
\begin{equation}
    \log p_\theta(x_0 \mid c, x_t) = s_\theta(x_0 \mid c, x_t) - \log Z(c, x_t).
\end{equation}
Since the normalization term is constant with respect to the token vocabulary $x_0$, we can directly manipulate the logits to approximate the operations on log-probabilities: $\log p_\theta \propto s_\theta$.

\paragraph{Decomposition as Product of Experts.}
Our goal is to construct a final guided distribution $\tilde{p}_\theta$ that simultaneously satisfies two objectives: (1) adhering to the problem constraints and (2) incorporating constructive refinements from the draft. We formulate this as the product of two distinct classifier-free guidance distributions:
\begin{equation}
    \tilde{p}_\theta(x_0) \propto \underbrace{\tilde{p}_{\text{problem}}(x_0)}_{\text{Problem Adherence}} \cdot \underbrace{\tilde{p}_{\text{refine}}(x_0)}_{\text{Refinement Direction}}.
\end{equation}

The first component, $\tilde{p}_{\text{problem}}$, applies standard CFG~\cite{cfg} to ensure the generation anchors to the problem statement $Q$ ($s_{prob}$) relative to the unconditional baseline $\emptyset$ ($s_{\emptyset}$):
\begin{equation}
    \tilde{p}_{\text{problem}}(x_0) \propto \frac{p_\theta(x_0 \mid Q)^{1+S_1}}{p_\theta(x_0 \mid \emptyset)^{S_1}}.
\end{equation}

The second component, $\tilde{p}_{\text{refine}}$, guides the generation toward the corrected logic found in the joint context $(Q, W)$ ($s_{joint}$) relative to the problem-only context ($s_{prob}$). This extracts the marginal improvement provided by the draft:
\begin{equation}
    \tilde{p}_{\text{refine}}(x_0) \propto \frac{p_\theta(x_0 \mid Q, W)^{1+S_2}}{p_\theta(x_0 \mid Q)^{S_2}}.
\end{equation}

\paragraph{Deriving the Logit Arithmetic.}
By substituting these definitions into the product formulation and taking the logarithm, we derive the additive logit update rule:
\begin{equation}
\begin{aligned}
    \log \tilde{p}_\theta(x_0) 
    = \log \tilde{p}_{\text{prob}}(x_0) + \log \tilde{p}_{\text{ref}}(x_0) + C \\
    = \big[ (1+S_1)\log p(Q) - S_1 \log p(\emptyset) \big] \\
     + \big[ (1+S_2)\log p(Q, W) - S_2 \log p(Q) \big].
\end{aligned}
\end{equation}
Rearranging the terms, we finally obtain:
\begin{equation}
\begin{aligned}
&\log \tilde{p}_\theta(x_0) \\
&= \big[ \log p(Q) + S_1(\log p(Q) - \log p(\emptyset)) \big] \\
& + \big[ \log p(Q,W)
    + S_2(\log p(Q,W) - \log p(Q)) \big].
\end{aligned}
\end{equation}
Replacing the log-probabilities with their corresponding logits ($s_{prob}$, $s_{\emptyset}$, $s_{joint}$), we can recover Eq.~\eqref{eq:final_update}:
\begin{equation}
\begin{aligned}
    \tilde{s}_\theta &= \underbrace{s_{prob} + S_1(s_{prob} - s_\emptyset)}_{\text{Problem Guidance}} \\
    &\quad + \underbrace{s_{joint} + S_2(s_{joint} - s_{prob})}_{\text{Refinement Guidance}}.
\end{aligned}
\end{equation}
This derivation confirms that our linear combination of logits is theoretically grounded in the intersection of two guided probability distributions.

\section{Experimental Details}
\label{app:exp_details}
In this section, we provide comprehensive details regarding the training configurations, model hyperparameters, and computational infrastructure used in our experiments.

\subsection{SFT Dataset Construction Details}
\label{app:data_details}

We curated a unified dataset balanced across three domains including Mathematics, Coding, and General Reasoning. To align the dataset size across domains, we targeted approximately 10,000 unique problems per domain which were then expanded into standard pairs $(Q, G)$ and refinement triplets $(Q, W, G)$. The final dataset contains 55,440 training samples and 2,913 validation samples. The detailed curation process for each domain is described below.

\paragraph{Mathematics (GSM8K, MATH).}We utilized the Math-Shepherd dataset~\cite{math-shepherd} as it contains explicit labels for incorrect reasoning steps. For each problem, we selected one incorrect solution to form the draft $W$ and paired it with the ground truth $G$. This process resulted in valid training pairs after length filtering with approximately 5,100 samples for GSM8K and 14,600 samples for MATH.

\paragraph{Coding (MBPP, Magicoder).}
We sourced problems from the MBPP training set and Magicoder. Since identifying incorrect code solutions requires computationally expensive execution-based testing, we employed a heuristic supervision strategy where we treated Llama-3.1-8B-Instruct generated solutions as potential drafts ($W$) without explicit execution verification. From the large-scale Magicoder dataset, we sampled 20,000 Python instances to match the volume of the mathematics domain. Combined with 474 MBPP problems, this yielded approximately 21,000 training samples in total.

\paragraph{General Reasoning (MMLU).}We used the MMLU-Auxiliary training set which contains 99.8k samples. To obtain hard negatives, we utilized Qwen3-8B to solve the training set and selectively collected failure cases where the model produced incorrect answers. These 12,055 failure instances served as the source for $W$. We sampled 10,000 unique problems from this subset and expanded them into pairs and triplets to create 20,000 training samples.

\paragraph{Preprocessing.}We applied a strict length filter to ensure training stability. Any sample where the combined prompt including the template and inputs exceeded 1,028 tokens was discarded. The final filtered dataset was split into training and validation sets using a stratified 19 to 1 ratio.

\begin{table*}[t]
    \centering
    \resizebox{0.85\textwidth}{!}{
    \begin{tabular}{l|c c | c c c | c}
        \toprule
        \multirow{3}{*}{\textbf{Benchmark}} & \multicolumn{2}{c|}{\textbf{LLM}} & \multicolumn{3}{c|}{\textbf{MDM Baselines (8B)}} & \textbf{Ours} \\
        \cmidrule(lr){2-3} \cmidrule(lr){3-6} \cmidrule(lr){7-7} 
         & \textbf{Self-refine} & \textbf{Self-refine} & \textbf{Standard} & \textbf{Dual-CFG} & \textbf{Standard} & \textbf{DCGC} \\
         &\small{Source(7B/32B)}  &\small{LLaMA-8B} & \small{LLaDA-} & \small{LLaDA} & \small{SFT LLaDA} & \small{SFT LLaDA} \\
        \midrule
        \multicolumn{7}{l}{\textit{\textbf{Source: Mistral-7B-v1}}} \\
        \midrule
        \textbf{GSM8K}       & 2.7  & 29.5  & \underline{35.3} & 30.4 & 6.7  & \textbf{42.0} \\
        \textbf{MATH}        & 2.1  & 1.7 & 11.0 & 16.4 & \underline{19.5} & \textbf{20.9} \\
        \textbf{MBPP}   & 25.9  & 34.8  & 0.0    & 11.9    & \underline{42.2}    & \textbf{43.8} \\
        \midrule
        \multicolumn{7}{l}{\textit{\textbf{Source: Qwen-2.5-32B}}} \\
        \midrule
        \textbf{GSM8K}       & 12.1 & 22.7 & 3.0  & 16.7 & \underline{19.7} & \textbf{28.8} \\
        \textbf{MATH}        & 0.6  & 0.0 & 12.8 & 10.3 & \underline{14.1} & \textbf{18.6} \\
        \textbf{MBPP}   & \textbf{29.4}    & 1.8    & 0.0 & 1.8    & 1.8    & \underline{8.3} \\
        \bottomrule
    \end{tabular}
    }
    \caption{\textbf{Generalizability across different source models.} Comparison of refinement performance on hard sets constructed by Mistral-7B-v1 and Qwen-2.5-32B. \textbf{SR} denotes Self-Refine.}
    \label{tab:generalizability_results}
\end{table*}

\subsection{Hard Set Construction Details}
\label{app:hardset}

To rigorously evaluate the refinement capability of our model, we constructed a \textbf{Hard Set} consisting exclusively of problems that the initial solver, \textbf{Llama-3.1-8B-Instruct}, failed to solve correctly. Specifically, we performed zero-shot inference on the training set of each benchmark and filtered out instances where the model's output matched the ground truth. The retained incorrect generations serve as the flawed solutions ($W$) for our refinement task.

The resulting Hard Set comprises the following number of samples: \textbf{GSM8K} (216), \textbf{MATH} (282), \textbf{MBPP} (216), \textbf{HumanEval} (69), \textbf{MMLU-STEM} (956), and \textbf{MMLU-Pro} (6,675). By focusing on these failure cases, we ensure that the evaluation strictly measures the model's ability to correct errors rather than its ability to solve easy problems from scratch.

\subsection{Baseline Details}
\label{app:baseline_details}
We implement the \textit{Self-Refine} baseline following \citet{selfrefine}, using the four prompt templates provided in Appendix~\ref{app:prompts-selfrefine}. To ensure reproducibility, we detail the configuration for autoregressive baselines. For inference, we use nucleus sampling with $p=0.9$. The temperature is set to 0.6 for standard generation, while for the \textit{Self-Refine} pipeline, we lower it to 0.3 during the feedback and refinement phases to encourage precise critique. The refinement process is limited to a single iteration to ensure fair comparison with DCGC, which does not rely on iterative refinement. We adjust the maximum token limits for the initial generation, feedback, and refinement stages based on domain complexity: 512, 384, and 512 tokens for mathematics and reasoning tasks (GSM8K, MATH, MMLU), and 800, 512, and 800 tokens for coding benchmarks (MBPP, HumanEval), respectively. For the fine-tuned Llama-3.1 baseline, we employ Low-Rank Adaptation (LoRA) with rank $r=16$ and $\alpha=32$. The model is trained for 3 epochs with a maximum sequence length of 2,048. We use a learning rate of $2 \times 10^{-5}$ with a cosine scheduler and a warmup ratio of 0.03.
%  The refinement process is performed with a fixed iteration count of 1. 

\subsection{Training Configuration}
We fine-tuned the LLaDA-8B-Instruct backbone using Supervised Fine-Tuning (SFT) with Low-Rank Adaptation (LoRA). To ensure parameter efficiency, LoRA was applied to all linear layers with a rank $r=32$, alpha $\alpha=64$, and a dropout rate of 0.05. The model was trained using the AdamW optimizer with a learning rate of $2 \times 10^{-5}$, employing a cosine learning rate scheduler with a warmup ratio of 0.1. The training process spanned 3 epochs with evaluations performed every 0.25 epochs. We set the maximum sequence length to 1,028 tokens and utilized bfloat16 precision. All experiments were conducted on 2 $\times$ NVIDIA A100 (80GB) GPUs, accelerated by DeepSpeed (ZeRO-2 stage) for memory optimization.

\subsection{Inference and Sampling}
\label{app:implementation_details}
For inference, we generate solutions using a masked diffusion process with 128 sampling steps ($T=128$). We adopted this sufficient number of steps to prioritize the precision of reasoning correction and to ensure robust denoising against the negative signals from wrong solutions.
The maximum generation length is set to 256 tokens for GSM8K, MATH, and Coding tasks, and extended to 512 tokens for MMLU benchmarks.

\paragraph{Hyperparameter Search and Robustness.}
To evaluate the robustness of our framework and prevent overfitting to specific datasets, we determined the hyperparameters $\alpha$ (structural scale) and $\beta$ (refinement scale) using only the GSM8K as a validation proxy. We performed a coarse grid search over $\{0.5, 1.0, 1.5\}$ and selected the configuration that yielded the best performance. These selected values ($\alpha=0.5, \beta=1.0$) were then kept fixed across all other benchmarks without further tuning. This "train-once, apply-everywhere" strategy for hyperparameters demonstrates the generalization capability of our Dynamic Dual-CFG mechanism.

\begin{table*}[t]
    \centering
    \footnotesize
    \resizebox{0.85\textwidth}{!}{%
    \begin{tabular}{lcccccc}
    \toprule
    \multirow{2}{*}{\textbf{Function}} & \multicolumn{2}{c}{\textbf{Math}} & \multicolumn{2}{c}{\textbf{Code}} & \textbf{Know.} & \multirow{2}{*}{\textbf{Avg.}} \\
    \cmidrule(lr){2-3} \cmidrule(lr){4-5} \cmidrule(lr){6-6}
     & \textbf{GSM8K} & \textbf{MATH} & \textbf{MBPP} & \textbf{HEval} & \textbf{STEM} & \\
    \midrule
    Tanh & \textbf{44.9} & 19.1 & 10.7 & 10.1 & \underline{35.4} & \underline{24.0} \\
    Sigmoid & 38.0 & \underline{19.9} & \textbf{11.6} & 8.7 & 34.7 & 22.6 \\
    Gated Tanh & \underline{42.1} & 19.2 & 10.7 & 13.0 & 34.7 & 23.9 \\
    Gated Sigmoid & 39.8 & 17.7 & \underline{11.1} & \textbf{14.5} & 33.3  &23.3  \\
    \textbf{ReLU (DCGC)} & \textbf{44.9} & \textbf{22.3} & 10.7 & \underline{13.1} & \textbf{35.7} & \textbf{25.3} \\
    \bottomrule
    \end{tabular}%
    }
    \caption{\textbf{Ablation on Scaling Functions.} Comparison of different activation functions for the refinement scaling factor ($S_2$). \textbf{ReLU} (Ours) achieves the highest average accuracy, effectively filtering out noise through its hard-gating mechanism, whereas Tanh and Sigmoid suffer from noise leakage on complex reasoning tasks.}
    \label{tab:activation_ablation}
\end{table*}

\section{Generalizability to Diverse Initial Solvers}
\label{app:various_hardset}

The primary objective of this experiment is to demonstrate that DCGC possesses a model-agnostic refinement capability and is not overfitted to the specific error patterns of the Llama-3 architecture used in training. Since our main evaluation relied exclusively on flawed solutions generated by Llama-3.1-8B-Instruct, there is a valid concern that the refinement performance might be biased toward the linguistic distinctiveness or specific failure modes of that particular generator. To address this and verify robustness, we extended our evaluation by constructing additional hard sets using two distinct LLMs including \textbf{Mistral-7B-v1} and \textbf{Qwen2.5-32B}.

These models were selected to represent a diverse range of reasoning capabilities and architectures. Mistral-7B serves as a representative of high-performance dense models in the 7B parameter class while Qwen2.5-32B represents a significantly larger and more capable model. We constructed the hard sets by collecting failure cases where these models generated incorrect solutions. The resulting datasets consist of 224 GSM8K, 292 MATH, and 495 MBPP samples for Mistral, and 66 GSM8K, 156 MATH, and 109 MBPP samples for Qwen. Note that the significantly smaller size of the Qwen-based hard set reflects its superior baseline reasoning capability.

\begin{table*}[t]
    \centering
    \resizebox{0.85\textwidth}{!}{
    \begin{tabular}{c l l c c c c c}
    \toprule
    SFT & Guidance & Scaling & GSM8K & MATH & MBPP & MMLU-STEM & Avg. \\
    \midrule
    -- & Standard Sampling & --   & 3.2  & 7.1  & 1.9  & 6.5  & 4.7  \\
    -- & Dynamic Dual-CFG  & ReLU & 19.4 & 13.8 & 9.7  & 17.2 & 15.0 \\
    \midrule
    \checkmark & Standard Sampling & --   & 27.8 & 15.3 & 7.4  & 19.1 & 17.4 \\
    \checkmark & \textbf{DCGC (Relative)} & \textbf{ReLU} 
    & \textbf{42.6} & \textbf{18.1} & \textbf{13.4} & \textbf{31.4} & \textbf{26.4} \\
    \bottomrule
    \end{tabular}}
    \caption{
    \textbf{Additional results on the DREAM backbone.}
    We evaluate DCGC on DREAM, another masked diffusion backbone, using the same solver-failure hard-set protocol as in Table~\ref{tab:main_results1}.
    The results show that the same trend observed with LLaDA also holds for DREAM: dual-capability SFT improves the base model, and DCGC further improves over the matched SFT backbone with standard sampling.
    These results are intended as supporting evidence for backbone transfer, while the primary controlled comparison remains Table~\ref{tab:main_results1}.
    }
    \label{tab:dream_results}
\end{table*}

Table~\ref{tab:generalizability_results} presents the performance of DCGC and baseline methods on these unseen error distributions. The results provide compelling evidence that the editing capabilities of DCGC are not confined to the specific error patterns of its training source. On the hard set constructed from Mistral-7B-v1, DCGC consistently outperforms all baselines across three benchmarks. Notably on GSM8K, DCGC achieves an accuracy of $42.0\%$ and markedly surpasses both the autoregressive Self-Refine baseline using LLaMA-8B and the standard SFT LLaDA baseline. This trend holds for the MATH and MBPP benchmarks as well where DCGC maintains its superiority and demonstrates its ability to effectively correct reasoning flaws generated by a different dense model architecture.

Furthermore, the results on the Qwen-2.5-32B hard set highlight the robustness of our framework even when applied to a significantly larger and more capable model. Despite the high difficulty of improving upon the 32B-scale drafts using an 8B refiner, DCGC successfully identifies and corrects errors that the original model failed to resolve. Specifically, DCGC achieves $28.8\%$ on GSM8K and $18.6\%$ on MATH. These results are particularly striking as they outperform the self-refinement capability of the much larger Qwen-32B model itself which only attained 12.1\% and 0.6\% on the respective tasks. These findings confirm that DCGC learns a generalized refinement prior rather than merely overfitting to the linguistic or logical idiosyncrasies of Llama-3 and thereby validates its potential as a universal refiner for diverse open-source LLMs.

\section{Impact of Different Scaling Functions}
\label{app:activation}

To validate our choice of the scaling function for the refinement term $S_2$, we compare our ReLU-based strategy against standard non-linear functions (Tanh, Sigmoid) and their gated variants designed to enforce non-negativity. Specifically, we define Gated Tanh as $\text{ReLU}(\tanh(x))$ and Gated Sigmoid as $2 \cdot \text{ReLU}(\sigma(x) - 0.5)$, where the shift and scaling ensure the output is exactly zero when the confidence gap is zero.
Table~\ref{tab:activation_ablation} shows that while the framework is generally effective across different functions, DCGC (ReLU) yields the most robust performance, particularly on logic-intensive benchmarks like MATH ($22.3$) and GSM8K ($44.9$).
We observe that standard Sigmoid underperforms on GSM8K ($38.0$), likely due to its "soft gating" behavior, which may allow minor noise from the flawed solution to affect the generation. Furthermore, Tanh-based variants, despite incorporating hard gating, lag behind ReLU in complex reasoning tasks. This suggests that the saturation property of Tanh may limit the guidance strength when the confidence gap is large, whereas the linear nature of ReLU allows for stronger signal injection proportional to the model's certainty. Based on these empirical observations, we adopt ReLU for its simplicity and effectiveness in balancing noise filtration and signal amplification.

\section{Generalization to the DREAM Backbone}
\label{app:dream_results}

Our main experiments use LLaDA-8B-Instruct as the masked diffusion backbone. 
To examine whether the observed gains are specific to this backbone, we additionally evaluate DCGC on DREAM, another masked diffusion model, under the same solver-failure hard-set protocol used in Table~\ref{tab:main_results1}. 
This experiment is intended as a supporting generalization check rather than the primary evidence for the method; the main controlled comparison remains the matched LLaDA experiment in Table~\ref{tab:main_results1}.

Table~\ref{tab:dream_results} shows that the overall trend transfers to DREAM. 
Without SFT, standard sampling achieves only 4.7 average accuracy, while applying Dynamic Dual-CFG to the base DREAM backbone improves the average to 15.0. 
After dual-capability SFT, standard sampling reaches 17.4 average accuracy, indicating that task-specific adaptation is also important for DREAM. 
Finally, combining SFT with relative dual guidance yields the strongest performance, with DCGC achieving 42.6 on GSM8K, 18.1 on MATH, 13.4 on MBPP, and 31.4 on MMLU-STEM.

These results suggest that the benefit of DCGC is not tied to a single diffusion backbone. 
In particular, DCGC improves over the matched DREAM-SFT standard sampling baseline by 14.8 points on GSM8K, 2.8 points on MATH, 6.0 points on MBPP, and 12.3 points on MMLU-STEM. 
The pattern is consistent with the LLaDA results: SFT strengthens the model's problem-only and draft-conditioned generation capabilities, while relative dual guidance further improves performance by modulating the draft-conditioned signal during global denoising.

We emphasize that this experiment does not claim universal transfer across all MDMs. 
Rather, it provides additional evidence that the proposed draft-conditioned global correction framework can be instantiated on more than one masked diffusion backbone.

% \section{Latency and GPU memory measurement}
% \label{app:latency}

\section{Additional Baselines}
\label{app:additional_baselines}
We further compare DCGC against recent tool-free critique-and-revise methods in the same regime: 
RCI~\cite{rci} and ReVISE~\cite{revise}, both evaluated under our hard-set protocol.

\begin{table}[h]
    \centering
    \caption{
    Comparison with tool-free, verifier-free revise methods under our hard-set protocol.
    }
    \label{tab:critique_revision}
    \resizebox{\columnwidth}{!}{%
    \begin{tabular}{lccccc}
        \toprule
        \textbf{Method}
        & \textbf{Tool-free}
        & \textbf{Verifier-free}
        & \textbf{GSM8K}
        & \textbf{MATH}
        & \textbf{MMLU-STEM} \\
        \midrule
        Self-Refine
        & \checkmark
        & \checkmark
        & 26.4
        & 11.0
        & 26.3 \\

        RCI~\citep{rci}
        & \checkmark
        & \checkmark
        & 19.4
        & 0.7
        & 29.9 \\

        ReVISE~\citep{revise}
        & \checkmark
        & \checkmark
        & 4.2
        & 0.0
        & 4.8 \\

        \textbf{DCGC}
        & \checkmark
        & \checkmark
        & \textbf{44.9}
        & \textbf{22.3}
        & \textbf{35.7} \\
        \bottomrule
    \end{tabular}%
    }
\end{table}

As shown in Table~\ref{tab:critique_revision}, consistent with prior findings that unaided self-critique yields limited and inconsistent gains on reasoning tasks, RCI and ReVISE provide only modest improvements on failed instances.
In contrast, DCGC's draft-conditioned global correction achieves substantially stronger performance across all three reasoning benchmarks.

\section{Empirical Analysis of Confidence-Guided Draft Reuse}
\label{app:confidence_gap_reuse}

We further analyze whether the relative confidence gap in DCGC serves its intended role of controlling draft reuse. Importantly, DCGC does not treat either absolute confidence or the relative gap as a verifier of logical correctness. At each position \(i\), it computes the relative confidence gap
\[
g^{(i)} = C_{\mathrm{joint}}^{(i)}
        - C_{\mathrm{prob}}^{(i)}
\]
and uses it to determine the residual scale
\[
S_2^{(i)} = \beta \cdot \operatorname{ReLU}
\left(g^{(i)}\right).
\]
Thus, additional draft-conditioned residual amplification is activated only when conditioning jointly on the problem and draft increases confidence relative to conditioning on the problem alone. The joint-conditioned logits remain part of the combined logits regardless of the gate; \(S_2\) controls only the additional amplification of the residual direction. This differs from relying on absolute joint confidence, as in the independent-scaling variant in Table~\ref{tab:main_results2}.

We evaluate the relationship between the confidence gap and draft reuse on MATH at both the token and example levels. Draft reuse is measured using 8-gram overlap between the generated solution and the input draft. For token-level reuse statistics, we consider gate-open positions, i.e., positions where \(S_2 > 0\).

\paragraph{Token-level analysis.}
We first divide token positions according to whether the problem-only and joint branches predict the same argmax token. A position is labeled as \textit{disagreement} when their argmax predictions differ and as
\textit{agreement} otherwise. Table~\ref{tab:token_gap_reuse} reports the mean confidence gap and draft-reuse rate for the two groups.

\begin{table}[t]
    \centering
    \small
    \setlength{\tabcolsep}{7pt}
    \begin{tabular}{lcc}
        \toprule
        \textbf{Branch relation}
        & \textbf{Mean gap}
        & \textbf{Draft reuse} \\
        \midrule
        Disagreement & 0.214 & 0.321 \\
        Agreement    & 0.026 & 0.192 \\
        \bottomrule
    \end{tabular}
    \caption{
    \textbf{Token-level relationship between branch disagreement,
    confidence gap, and draft reuse on MATH.}
    Disagreement indicates positions at which the problem-only and
    joint branches predict different argmax tokens. Draft reuse is
    measured by 8-gram overlap at gate-open positions.
    }
    \label{tab:token_gap_reuse}
\end{table}

The mean confidence gap is approximately eight times larger at disagreement positions than at agreement positions (\(0.214\) vs.\ \(0.026\)). Draft reuse is also substantially higher at disagreement positions (\(0.321\) vs.\ \(0.192\)). Moreover, the gate scale \(S_2\) increases monotonically with branch-disagreement
probability, with a Spearman correlation of \(\rho=0.167\) and a 95\% confidence interval of \([0.150, 0.182]\). These results indicate that the relative gap concentrates additional residual amplification at positions where the two conditioning branches provide competing predictions. At agreement positions, the gap and corresponding amplification are considerably smaller, although not necessarily zero.

\paragraph{Example-level analysis.}
We next test whether confidence-guided reuse reflects the relevance of the draft to the current problem rather than superficial factors such as generation length or the mere presence of draft context. For each
example, we compute the mean confidence gap and its draft-reuse rate. We then measure their Spearman correlation under two conditions: the original draft paired with its corresponding problem and a
shuffled control in which the problem is paired with a draft from an unrelated MATH example. We additionally report partial correlations controlling for generation length.

\begin{table}[t]
    \centering
    \small
    \setlength{\tabcolsep}{5pt}
    \resizebox{\columnwidth}{!}{%
    \begin{tabular}{lccc}
        \toprule
        \textbf{Draft}
        & \(\boldsymbol{\rho}\)
        & \textbf{95\% CI}
        & \(\boldsymbol{\rho}\) \textbf{(length-controlled)} \\
        \midrule
        Original
        & \(+0.195\)
        & \([+0.070, +0.314]\)
        & \(+0.247\) \\
        Shuffled
        & \(+0.078\)
        & \([-0.051, +0.204]\)
        & \(+0.077\) \\
        \bottomrule
    \end{tabular}%
    }
    \caption{
    \textbf{Example-level correlation between the mean confidence gap and draft reuse on MATH.} The shuffled condition pairs each problem with a draft from an unrelated example. The final column reports the partial Spearman correlation after controlling for generation length.}
    \label{tab:example_gap_reuse}
\end{table}

With the original drafts, the confidence gap is significantly correlated with draft reuse (\(\rho=0.195\)), and the correlation remains positive after controlling for generation length (\(\rho=0.247\)). In contrast, the correlation is substantially weaker under the shuffled control, and its confidence interval includes zero.
Shuffling preserves the presence of a draft while removing its semantic correspondence with the problem. The resulting difference therefore suggests that the confidence gap responds to problem--draft alignment rather than merely to draft availability or output length.

Together, the token- and example-level analyses support the intended mechanistic role of the relative confidence gap: it controls where and how strongly draft-derived information is reused. These results do not
imply that the gap verifies logical correctness or that a larger gap necessarily guarantees successful correction. Instead, they show that the gap functions as a selective draft-reuse signal rather than an indiscriminate confidence heuristic.

\section{Prompts}
\label{app:prompts}

% \subsection{Hardset Prompt Template}
% \label{app:prompts-hardset}

\subsection{Self-Refine Prompt Format}
\label{app:prompts-selfrefine}
We provide the four prompt templates used in the \textit{Self-Refine} pipeline, which respectively guide initial solution generation, self-critique, refinement, and final response selection. 

\begin{tcolorbox}[title=\textbf{Generation Prompt}, breakable, listing only, listing options={style=promptstyle}]
You are a careful math tutor. Solve the problem step by step. Follow below final answer format strictly!
Keep each step short and factual. At the very end, output exactly one line:
The answer is: <number>
\par\smallskip
Problem: \textcolor{NavyBlue}{[Problem here]}

\par\smallskip
Solution:
\end{tcolorbox}

\begin{tcolorbox}[title=\textbf{Feedback Prompt}, breakable, listing only, listing options={style=promptstyle}]
You are a strict reviewer. Given the math problem and the current solution, write specific, actionable feedback listing concrete fixes.
Point out wrong arithmetic, missing constraints, or invalid reasoning.
If the solution is already fully correct and clearly presented, say so.
\par\smallskip
Problem:
\textcolor{NavyBlue}{[Problem here]}
\par\smallskip
Current solution:
\textcolor{NavyBlue}{[Draft Solution here]}
\par\smallskip
Feedback (bulleted list of fixes and checks):

Finally, output exactly one line as 'Stop: yes' if no further refinement is needed, otherwise 'Stop: no'. Do not output anything after this line.
Stop:
\end{tcolorbox}

\begin{tcolorbox}[title=\textbf{Refinement Prompt}, breakable, listing only, listing options={style=promptstyle}]
You are an expert math editor. Improve the solution based on the feedback.
Fix every pointed issue. Keep the reasoning concise and correct.
At the very end, output exactly one line:
The answer is: <number>
\par\smallskip
Problem:
\textcolor{NavyBlue}{[Problem here]}
\par\smallskip
Previous solution:
\textcolor{NavyBlue}{[Draft Solution here]}
\par\smallskip
Feedback:
\textcolor{NavyBlue}{[Feedback here]}
\par\smallskip
Improved solution:
\end{tcolorbox}

\begin{tcolorbox}[title=\textbf{System Prompt / Instruction Header (Optional)}, breakable, listing only, listing options={style=promptstyle}]
You are a helpful assistant. Follow the instructions precisely and output only what is requested.
\end{tcolorbox}

\subsection{DCGC Prompt Template}
\label{app:prompts-DCGC}

\begin{tcolorbox}[
    title=\textbf{Standard Solving prompts for Problem Context}, 
    % colback=white, 
    % colframe=black,
    % fontupper=\small, % 폰트 크기 조절 (필요시)
    % boxrule=1pt,
    % arc=2mm
    % fontupper=\rmfamily
]
    \textbf{\textit{Mathematics (GSM8K, MATH)}}
    \par\smallskip
    You are an expert mathematician. Solve the problem step-by-step, showing your rigorous reasoning and calculations. End with "The answer is: X" where X is your final solution.
    \par\smallskip
    Question: \textcolor{NavyBlue}{[Problem here]}
    \par\bigskip
    
    \textbf{\textit{Coding (MBPP, HumanEval)}}
    \par\smallskip
    You are an expert programming assistant. Write a correct and efficient solution to the following coding problem. Provide your reasoning if necessary, and output the code inside a markdown block.
    \par\smallskip
    Problem: \textcolor{NavyBlue}{[Problem here]}
    \par\bigskip

    \textbf{\textit{General Knowledge (MMLU)}}
    \par\smallskip
    Solve the following multiple-choice question step-by-step. Think through the problem logically, show your reasoning chain, and justify your choice with specific evidence or calculations. Ensure your reasoning leads directly to the correct option. End your response with: "The answer is: (X)" where X is the correct choice (A, B, C, or D).
    \par\smallskip
    Question: \textcolor{NavyBlue}{[Question here]}
\end{tcolorbox}

\begin{tcolorbox}[title=\textbf{Refinement prompts for Joint Context}, breakable, listing only, listing options={style=promptstyle}]
You are an expert mathematician. Solve the problem step-by-step, showing your rigorous reasoning and calculations. End with "The answer is: X" where X is your final solution.
\par\smallskip
Question: \textcolor{NavyBlue}{[Problem here]}
\par\smallskip
\textcolor{NavyBlue}{[Draft Solution here]}

Using the provided solution above as a reference, derive a rigorous, correct step-by-step solution. Verify the logic and calculations within the reference, correcting any inaccuracies only when necessary to ensure the final answer is precise. 
% Ensure the final output follows the required format.
\end{tcolorbox}

\section{Qualitative Analysis}
\label{app-sec:qual}

\subsection{Step-by-step Refinement}
Table~\ref{app-tab:qualitative_example} provides a step-by-step visualization of the DCGC generation process across diffusion timesteps. Rather than indiscriminately following the flawed initial draft, DCGC dynamically modulates its guidance based on the refinement confidence. For instance, at early stages (e.g., $t=7$), the model effectively filters out the draft's incorrect logic by adhering to the problem-conditioned structure ($C_{prob}$). Conversely, at $t=20$ and $t=33$, when the joint-conditioned confidence is sufficiently high ($C_{joint} > C_{prob}$), DCGC selectively injects constructive math hints to correct the reasoning trajectory. Through this token-level adaptation, the framework successfully rectifies the wrong draft into the correct final answer.

\begin{table*}[t]
\centering
\small
\renewcommand{\arraystretch}{1.3} % 행 간격 조절
% \begin{tabular}{p{0.9cm} p{3.7cm} p{3.7cm} p{3.7cm} p{3cm}}
% \toprule
\begin{tabular}{|p{15.4cm}|}
\hline
\textbf{Q:} Claire makes a 3 egg omelet every morning for breakfast. How many dozens of eggs will she eat in 4 weeks? \\
\textbf{W:} ...4 weeks * 7 days/week = 28 days...3 omelets/day * 28 days = 84 ...84 omelets * 3 eggs/omelet = 252 eggs... The answer is: 21 \\
\hline
\end{tabular}

\vspace{0.2cm}

\begin{tabular}{p{0.6cm}|p{3.3cm}|p{3.3cm}|p{3.3cm}|p{3.0cm}}
\hline
\textbf{Step} & \textbf{Unconditional} & \textbf{Problem-Conditioned} & \textbf{Joint-Conditioned} & \textbf{DCGC Action} \\
\hline

% Step 7
$t=7$ & 
...3 eggs ... for \textcolor{errorred}{\textbf{5 days}} ... \textcolor{errorred}{\textbf{3*5}} & 
... 3 eggs ... for \textcolor{correctblue}{\textbf{7 days}}... \textcolor{correctblue}{\textbf{3*7}} & 
... 3 eggs ... for \textcolor{correctblue}{\textbf{7 days}} ... \textcolor{errorred}{\textbf{3*3}}  & 
\textbf{Follow Structure} ($C_{prob}$)  \\ 
\hline

% Step 19
$t=19$ & 
...so she eats {\textbf{3*7 = 21}} eggs... & 
...so she eats {\textbf{3*7 = 21}} eggs... & 
...so she eats {\textbf{3*7 = 21}} eggs... & 
Consensus  \\ 
\hline

% Step 20
$t=20$ & 
Over 4 weeks ... \textcolor{errorred}{\textbf{2 * 4}}... & 
Over 4 weeks ... \textcolor{errorred}{\textbf{1 * 4}}... & 
Over 4 weeks ... \textcolor{correctblue}{\textbf{21 * 4 = 84}}... & 
\textbf{Inject Hint} ($C_{joint} > C_{prob}$) \\ 
\hline

% Step 29
$t=29$ & 
...will eats {\textbf{21 * 4 = 84}} eggs & 
...will eat {\textbf{21 * 4 = 84}} eggs & 
...will eat {\textbf{21 * 4 = 84}} eggs & 
Consensus  \\ 
\hline

% Step 33
$t=33$ & 
A dozen is 12 eggs so she needs \textcolor{errorred}{\textbf{84///8}}... & 
A dozen is 12 eggs so she will \textcolor{errorred}{\textbf{84//1}}... & 
A dozen is 12 eggs so she eats \textcolor{correctblue}{\textbf{84/12}}... & 
\textbf{Inject Hint} ($C_{joint} > C_{prob}$)  \\ 
\hline

% Step 38
$t=38$ & 
...so she eats {\textbf{84/12 = 7}} dozens & 
...so she eats {\textbf{84/12 = 7}} dozens & 
...so she eats {\textbf{84/12 = 7}} dozens & 
Consensus  \\ 
\hline

% Step 47
$t=47$ & 
The answer is \textcolor{correctblue}{\textbf{7}} & 
The answer is: \textcolor{correctblue}{\textbf{7}} & 
The answer is: \textcolor{correctblue}{\textbf{7}} & 
  \\ 

\bottomrule
\end{tabular}
\caption{\textbf{Step-by-step visualization of the DCGC generation process on a GSM8K example.} The table illustrates how dynamic dual-CFG adapts at the token level. DCGC effectively filters out incorrect signals from the wrong solution (e.g., $t=7$) while selectively injecting constructive hints (e.g., $t=20, 33$) only when the refinement context provides higher confidence ($C_2 > C_1$). Red and blue text indicate incorrect and corrected tokens, respectively.}
\label{app-tab:qualitative_example}
\end{table*}

\subsection{Case Studies: Error Correction across Benchmarks}
We provide concrete qualitative examples to illustrate how DCGC successfully rectifies flawed initial solutions compared to baseline refinement methods, as shown in Figure~\ref{app-fig:qualitative_river_full} (GSM8K) and Figure~\ref{app-fig:qualitative_integer_power_full} (MATH). 
A critical limitation of ARM-based refinement methods is error propagation. In contrast, the parallel decoding mechanism of our MDM-based DCGC framework evaluates and updates the entire sequence simultaneously. This global context awareness enables DCGC to safely filter out logical noise from the draft while injecting necessary corrections at the precise locations required.

\begin{figure*}[ht]
    \centering
    \small
    \renewcommand{\arraystretch}{1.1}
    \begin{tcolorbox}[colback=white, colframe=gray!60, title=\textbf{Qualitative Comparison on GSM8K}, fonttitle=\bfseries, boxrule=0.8pt, arc=2mm, left=2mm, right=2mm, top=2mm, bottom=2mm]
        
        % --- Question ---
        \textbf{Question:} The amount of water passing through a river at one point in time is 4000 gallons. After a day of heavy rain, the amount of water passing through the river doubles at the same point. If the volume of water passing through the river at that point increases by 6000 gallons on the third day, calculate the \underline{total amount} of water passing through the river at that point.
        \par\noindent\rule{\textwidth}{0.4pt} \vspace{0.1cm}

        % --- Wrong Solution (Draft) ---
        \textbf{Wrong Solution (Draft $W$)} \\
        Step 1: The initial amount of water passing through the river is 4000 gallons. \\
        Step 2: After a day of heavy rain, the amount of water doubles, so it becomes $4000 * 2 = 8000$ gallons. \\
        Step 3: On the third day, the volume of water increases by 6000 gallons, so the new amount is $8000 + 6000 = 14000$ gallons. \\
        \textcolor{red}{Step 4: To find the total amount of water passing through the river at that point, we need to add the initial amount to the amount on the third day, which is $4000 + 14000 = 16000$ gallons.} \\
        \textbf{\textcolor{red}{The answer is: 16000}}
        \par\noindent\rule{\textwidth}{0.1pt} \vspace{0.1cm}

        % --- Self-Refine ---
        \textbf{Self-Refine (Baseline)} \\
        Step 1: The initial amount of water passing through the river is 4000 gallons. \\
        Step 2: After a day of heavy rain, the amount of water doubles, so it becomes $4000 * 2 = 8000$ gallons. \\
        Step 3: On the second day, the amount of water remains 8000 gallons. \\
        Step 4: On the third day, the volume of water increases by 6000 gallons, so the new amount is $8000 + 6000 = 14000$ gallons. \\
        \textcolor{red}{Step 5: To find the total amount of water passing through the river at that point, we need to add the initial amount, the amount on the second day, and the amount on the third day, which is $4000 + 8000 + 14000 = 24000$ gallons.} \\
        \textbf{\textcolor{red}{The answer is: 24000}}
        \par\noindent\rule{\textwidth}{0.1pt} \vspace{0.1cm}

        % --- Standard Sampling ---
        \textbf{$\text{LLaDA}_{\text{SFT}}$ Standard Sampling (Baseline)} \\
        Step 1: The initial amount of water passing through the river is 4000 gallons. \\
        Step 2: After a day of heavy rain, the amount of water doubles, so it becomes $4000 * 2 = 8000$ gallons. \\
        Step 3: On the third day, the volume of water increases by 6000 gallons, so the new amount is $8000 + 6000 = 14000$ gallons. \\
        \textcolor{red}{Step 4: To find the total amount of water passing through the river at that point, we need to add the initial amount to the amount on the third day, which is $4000 + 14000 = 18000$ gallons.} \\
        \textbf{\textcolor{red}{The answer is: 18000}}
        \par\noindent\rule{\textwidth}{0.1pt} \vspace{0.1cm}

        % --- DCGC (Ours) ---
        \textbf{DCGC (Ours)} \\
        After a day of heavy rain, the amount of water passing through the river doubled to $4000*2=8000$ gallons. \\
        \textcolor{blue}{On the third day, the volume of water passing through the river increased by 6000 gallons to totaling $8000+6000=14000$ gallons.} \\
        \textbf{\textcolor{blue}{The answer is: 14000}}
        
    \end{tcolorbox}
    \caption{\textbf{Qualitative Example on GSM8K.} The phrase "total amount" misleads the Draft into incorrectly summing past states ($4000+14000$). \textbf{Self-Refine} exacerbates this by summing all days. \textbf{Standard Sampling} copies the Draft's flawed logic verbatim (Step 4) but corrects the arithmetic error ($16000 \to 18000$). In contrast, \textbf{DCGC} correctly interprets the final state volume, filtering out the logical noise.}
    \label{app-fig:qualitative_river_full}
\end{figure*}

\begin{figure*}[ht]
    \centering
    \small
    \renewcommand{\arraystretch}{1.1}
    % 필요한 패키지: \usepackage{tcolorbox}, \usepackage{xcolor}
    \begin{tcolorbox}[colback=white, colframe=gray!60, title=\textbf{Qualitative Comparison on MATH}, fonttitle=\bfseries, boxrule=0.8pt, arc=2mm, left=2mm, right=2mm, top=2mm, bottom=2mm]
        
        % --- Question ---
        \textbf{Question:} For how many positive integers $n>1$ is it true that $2^{24}$ is a perfect $n^{\text{th}}$ power?
        \par\noindent\rule{\textwidth}{0.4pt} \vspace{0.1cm}

        % --- Wrong Solution (Draft) ---
        \textbf{Wrong Solution (Draft $W$)} \\
        Step 1: To solve this problem, we need to find the positive integers $n>1$ for which $2^{24}$ is a perfect $n^{\text{th}}$ power. \\
        Step 2: We start by expressing $2^{24}$ as a product of its prime factors. Since $2^{24} = (2^4)^6$, we can rewrite it as $16^6$. \\
        Step 3: This means that $2^{24}$ can be expressed as a perfect $6^{\text{th}}$ power, as $16 = 2^4$. \\
        Step 4: Now, we need to find other possible values of $n$ for which $2^{24}$ is a perfect $n^{\text{th}}$ power. \\
        Step 5: We can express $2^{24}$ as $(2^2)^{12}$, which equals $4^{12}$. This implies that $2^{24}$ is also a perfect $12^{\text{th}}$ power. \\
        Step 6: Additionally, we can express $2^{24}$ as $(2^3)^8$, which equals $8^8$. This shows that $2^{24}$ is also a perfect $8^{\text{th}}$ power. \\
        \textcolor{red}{Step 7: However, we cannot express $2^{24}$ as a perfect $n^{\text{th}}$ power for any other value of $n$ greater than 8, as the prime factorization of $2^{24}$ is $2^{24}$ itself, and no other perfect power of $n$ can have a prime factorization with only 2s.} \\
        \textcolor{red}{Step 8: Therefore, the positive integers $n>1$ for which $2^{24}$ is a perfect $n^{\text{th}}$ power are 6, 8, and 12.} \\
        \textbf{\textcolor{red}{The answer is: 3}}
        \par\noindent\rule{\textwidth}{0.1pt} \vspace{0.1cm}

        % --- Self-Refine ---
        \textbf{Self-Refine (Baseline)} \\
        Step 1: To solve this problem, we need to find the positive integers $n>1$ for which $2^{24}$ is a perfect $n^{\text{th}}$ power. \\
        Step 2: We start by expressing $2^{24}$ as a product of its prime factors. Since $2^{24} = (2^4)^6$, we can rewrite it as $16^6$. This is because $16 = 2^4$, and therefore, $16^6 = (2^4)^6 = 2^{24}$. \\
        Step 3: This means that $2^{24}$ can be expressed as a perfect $6^{\text{th}}$ power, as $16^6 = 2^{24}$. \\
        Step 4: Now, we need to find other possible values of $n$ for which $2^{24}$ is a perfect $n^{\text{th}}$ power. \\
        Step 5: We can express $2^{24}$ as $(2^2)^{12}$, which equals $4^{12}$. This is because $4 = 2^2$, and therefore, $4^{12} = (2^2)^{12} = 2^{24}$. \\
        Step 6: Additionally, we can express $2^{24}$ as $(2^3)^8$, which equals $8^8$. This is because $8 = 2^3$, and therefore, $8^8 = (2^3)^8 = 2^{24}$. \\
        \textcolor{red}{Step 7: However, we cannot express $2^{24}$ as a perfect $n^{\text{th}}$ power for any other value of $n$ greater than 8, as the prime factorization of $2^{24}$ is $2^{24}$ itself, and no other perfect power of $n$ can have a prime factorization with only 2s.} \\
        Step 8: Therefore, the positive integers $n>1$ for which $2^{24}$ is a perfect $n^{\text{th}}$ power are 6, 8, and 12. \\
        \textbf{\textcolor{red}{The answer is: 3}}
        \par\noindent\rule{\textwidth}{0.1pt} \vspace{0.1cm}

        % --- Standard Sampling ---
        \textbf{$\text{LLaDA}_{\text{SFT}}$ Standard Sampling (Baseline)} \\
        If $2^{24}$ is a perfect $n^{\text{th}}$ power, then $2^{24}=a^n$ for some integer $a$. Since $2^{24}=(2^3)^8$, $2^{24}=(2^4)^6$, and $2^{24}=(2^2)^{12}$, the values of $n$ are 6, 8, and 12. \textcolor{red}{Thus are only 3 such integers.} \\
        \textbf{\textcolor{red}{The answer is: 3}}
        \par\noindent\rule{\textwidth}{0.1pt} \vspace{0.1cm}

        % --- DCGC (Ours) ---
        \textbf{DCGC (Ours)} \\
        If $2^{24}$ is a perfect $n^{\text{th}}$ power, then $2^{24} = k^n$ for some integer $k$. \textcolor{blue}{Since $2^{24} = (2^4)^6 = (2^3)^8 = (2^2)^{12}$, we see that $n$ must be a divisor of 24.} The divisors of 24 are 1, 2, 3, 4, 6, 8, 12, and 24. However, we are only interested in the divisors greater than 1, so we have 2, 3, 4, 6, 8, 12, and 24. \textcolor{blue}{Thus, there are 7 such integers.} \\
        \textbf{\textcolor{blue}{The answer is: 7}}
        
    \end{tcolorbox}
    \caption{\textbf{Qualitative Example on MATH.} The Draft finds specific instances ($n=6, 8, 12$) but incorrectly concludes these are the \textit{only} solutions (Step 7). \textbf{Self-Refine} repeats this limited reasoning verbatim. \textbf{Standard Sampling} also exhibits tunnel vision. In contrast, \textbf{DCGC} generalizes from the specific examples to identify the underlying rule ($n$ is a divisor of 24), successfully finding all 7 solutions.}
    \label{app-fig:qualitative_integer_power_full}
\end{figure*}

\end{document}